%% file: _main.tex
\pdfoutput=1
\documentclass[sn-mathphys,Numbered]{sn-jnl}

\usepackage{graphicx}%
\usepackage{multirow}%
\usepackage{amsmath,amssymb,amsfonts}%
\usepackage{amsthm}%
\usepackage{mathrsfs}%
\usepackage[title]{appendix}%
\usepackage{xcolor}%
\usepackage{textcomp}%
\usepackage{manyfoot}%
\usepackage{booktabs}%
\usepackage{listings}%

\usepackage{amsmath}
\usepackage{amssymb}
\usepackage{mathtools}
\usepackage{amsthm}

\input{preamble.tex}

\theoremstyle{plain}

\theoremstyle{definition}

\theoremstyle{remark}

\begin{document}

\title{Augmenting Interpretable Models with LLMs during Training}

\author[1]{\fnm{Chandan} \sur{Singh}}\email{chansingh@microsoft.com}

\author[2]{\fnm{Armin} \sur{Askari}}\email{aaskari@berkeley.edu}

\author[1]{\fnm{Rich} \sur{Caruana}}\email{rcaruana@microsoft.com}

\author[1]{\fnm{Jianfeng} \sur{Gao}}\email{jfgao@microsoft.com}

\affil[1]{\orgdiv{Microsoft Research}} 

\affil[2]{\orgdiv{University of California, Berkeley}} 

\abstract{
Recent large language models (LLMs) have demonstrated remarkable prediction performance for a growing array of tasks.
However, their proliferation into high-stakes domains (e.g. medicine) and compute-limited settings has created a burgeoning need for interpretability and efficiency.
We address this need by proposing \methodmainlongs (\methodmain), a framework for leveraging the knowledge learned by LLMs to build extremely efficient and interpretable models.
\methodmains use LLMs during fitting but not during inference, allowing complete transparency and often a speed/memory improvement of greater than 1,000x for inference compared to LLMs.
We explore two instantiations of \methodmains in natural-language processing: (i) \method, which augments a generalized additive model with decoupled embeddings from an LLM and (ii) \methodt, which augments a decision tree with LLM feature expansions.
Across a variety of text-classification datasets, both outperform their non-augmented counterparts.
\methods can even outperform much larger models (e.g. a 6-billion parameter GPT-J model), despite having 10,000x fewer parameters and being fully transparent.
We further explore \methodmains in a natural-language fMRI study, where they generate interesting interpretations from scientific data.
All code for using \methodmains and reproducing results is made available on Github.\footnote{Scikit-learn-compatible API available at \href{https://github.com/csinva/imodelsX}{\faGithub~github.com/csinva/imodelsX} and experiments code available at \href{https://github.com/microsoft/augmented-interpretable-models}{\faGithub~github.com/microsoft/augmented-interpretable-models}.}
}

\maketitle

\section{Introduction}

Large language models (LLMs) have demonstrated remarkable predictive performance across a growing range of diverse tasks~\cite{brown2020language,bubeck2023sparks,devlin2018bert}.
However, their proliferation has led to two burgeoning problems.
First, like most deep neural nets, LLMs have become increasingly difficult to interpret,
often leading to them being characterized as black boxes and debilitating their use in high-stakes applications such as science~\cite{angermueller2016deep}, medicine~\cite{Kornblith2022}, and policy-making~\cite{brennan2013emergence}.
Moreover, the use of black-box models such as LLMs has come under increasing scrutiny in settings where users require explanations or where models struggle with issues such as fairness \cite{dwork2012fairness} and regulatory pressure \cite{goodman2016european}.
Second, black-box LLMs have grown to massive sizes, incurring enormous energy costs~\cite{bommasani2023ecosystem} and making them costly and difficult to deploy, particularly in low-compute settings (e.g. edge devices).

As an alternative to large black-box models, transparent models, such as generalized additive models~\cite{hastie1986generalized} and decision trees~\cite{breiman1984classification} can maintain complete interpretability.
Additionally, transparent models tend to be dramatically more computationally efficient than LLMs.
While transparent models can sometimes perform as well as black-box LLMs~\cite{rudin2021interpretable,ha2021adaptive,mignan2019one,tan2022Fast},
in many settings (such as natural-language processing (NLP)), there is often a considerable gap between the performance of transparent models and black-box LLMs.

\begin{figure}[ht]
    \centering
    \includegraphics[width=0.85\textwidth]{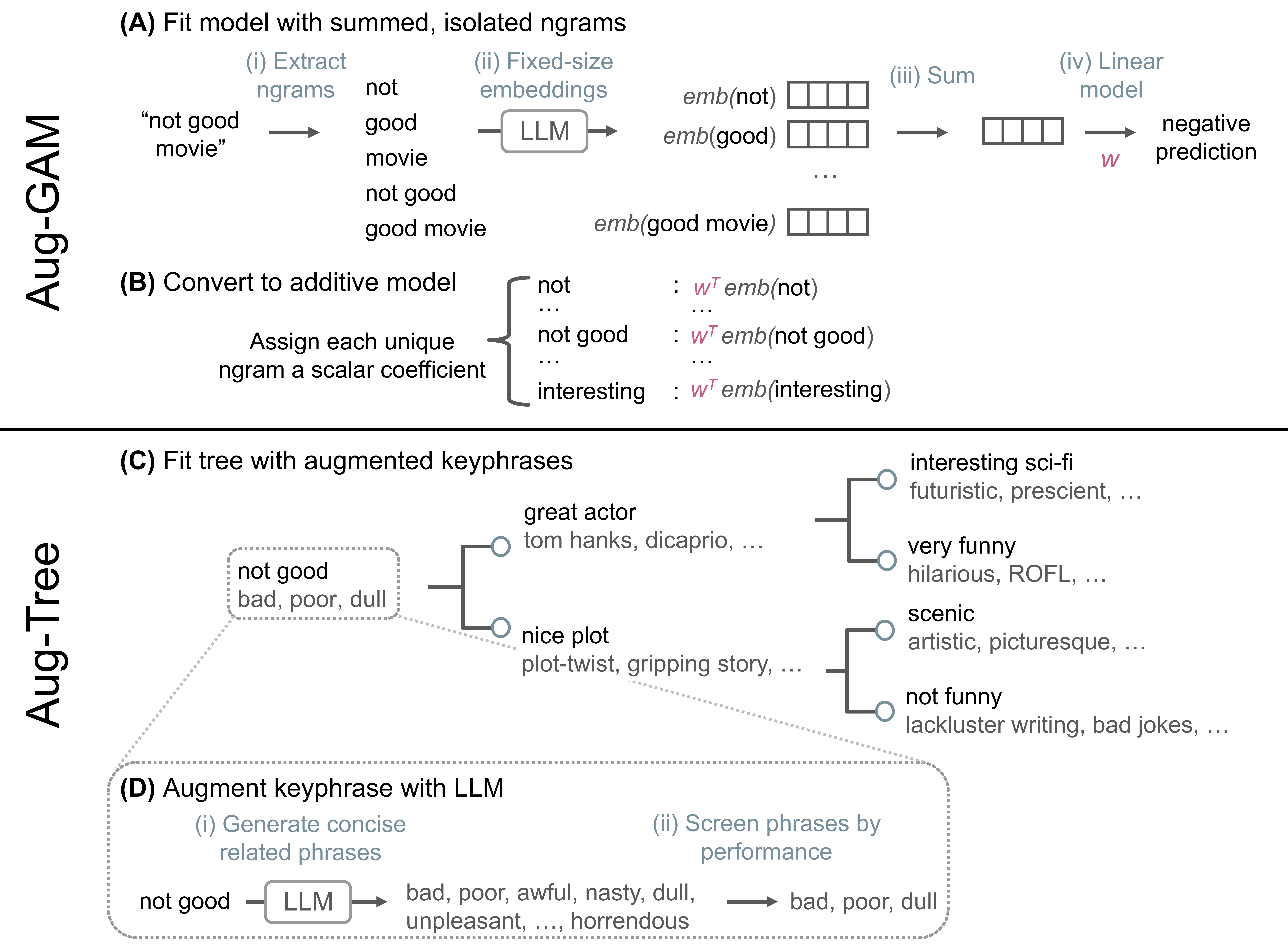}
    \caption{
    \methodmains use an LLM to augment an interpretable model during fitting but not inference (toy example for movie-review classification).
    \textbf{(A)} \methods fits an augmented additive model by
    extracting fixed-size embeddings for decoupled ngrams in a given sequence, summing them, and using them to train a supervised linear model.
    \textbf{(B)} At test time, \methods can be interpreted exactly as a generalized additive model.
    A linear coefficient for each ngram in the input is obtained by taking the dot product between the ngram's embedding and the shared vector $w$.
    \textbf{(C)} \methodts improves each split of a decision tree during fitting by
    \textbf{(D)} augmenting each keyphrase found by CART with similar keyphrases generated by an LLM. 
    }
    \label{fig:intro}
\end{figure}

We address this gap by proposing \methodmainlongs (\methodmain), a framework to leverage the knowledge learned by LLMs to build extremely efficient and interpretable models.
 Specifically, we define an \textbf{\methodmainsingular} as a method that leverages an LLM to fit an interpretable model, but \emph{does not use the LLM during inference}.
 This allows complete transparency and often a substantial efficiency improvement (both in terms of speed and memory).
\methodmains can address shortcomings in existing transparent models by using the world knowledge present in modern LLMs, such as information about feature correlations.

We explore two instantiations of \methodmain: (i) \method, which augments a generalized additive model via decoupled embeddings from an LLM and (ii) \methodt, which augments a decision tree with improved features generated by calling an LLM (see \cref{fig:intro}).
At inference time, both are completely transparent and efficient:
\methods requires only summing coefficients from a fixed dictionary 
while \methodts requires checking for the presence of keyphrases in an input.

Across a variety of natural-language-processing datasets, both proposed \methodmains outperform their non-augmented counterparts.
\methods can even outperform much larger models (e.g. a 6-billion parameter GPT-J model~\cite{gpt_j}), despite having 10,000x fewer parameters and no nonlinearities.
We further explore \methodmains in a natural-language fMRI context, where we find that they can predict well and generate interesting interpretations.
In what follows,
\cref{sec:methods} formally introduces \methodmain,
\cref{sec:results} and \cref{sec:interpretation} shows results for predictive performance and interpretation,
\cref{sec:fMRI} explores \methodmains in an fMRI prediction setting,
\cref{sec:background} reviews related work, 
and \cref{sec:discussion} concludes with a discussion.

\section{\methodmains methodology: \methods and \methodts}
\label{sec:methods}

In this section, \cref{subsec:limitations} overviews limitations of existing transparent methods,
\cref{subsec:gam_methods} introduces \method,
and \cref{subsec:tree_methods} introduces \methodt.

\subsection{Limitations of existing transparent methods}
\label{subsec:limitations}

We are given a dataset of $n$ natural language strings $X_{\text{text}}$ and corresponding labels $\by \in \mathbb R^n$.
In transparent modeling,
often each string $x$ is represented by a bag-of-words,
in which each feature $x_i$ is a binary indicator (or count) of the presence of a single token (e.g. the word \textit{good}).
To model interactions between tokens,
one can instead use a bag-of-ngrams representation, whereby 
each feature is formed by concatenating multiple tokens (e.g. the phrase \textit{not good}).
Using a bag-of-ngrams representation maps $X_{\text{text}}$ to a feature matrix $X \in \mathbb R^{n \times p}$, where $p$ is the number of unique ngrams in $X_{\text{text}}$.
While this representation enables interpretability, the number of ngrams in a dataset grows exponentially with the size of the ngram (how many tokens it contains) and the vocab-size; even for a modest vocab-size of 10,000 tokens,
the number of possible trigrams is already $10^{12}$.
This makes it
difficult for existing transparent methods to model all trigrams without overfitting.
Moreover, existing transparent methods 
completely fail to learn about ngrams not seen in the training set.

\paragraph{Preliminaries: GAMs}
Generalized additive models, or GAMs~\cite{hastie1986generalized} take the form:

\begin{align}
    \label{eq:GAM}
    g(\mathbb{E}[y])=\beta+f_1\left(x_1\right)+f_2\left(x_2\right)+\cdots+f_K\left(x_p\right),
\end{align}

where $\left(x_1, x_2, \ldots, x_p\right)$ are the input features (i.e. ngrams),
$y$ is the target variable, $g(\cdot)$ is the link function (e.g., logistic function) and each $f_i$ is a univariate shape function with $\mathbb{E}\left[f_i\right]=0$.
Due to the function's additivity, each component function $f_i$ can be interpreted independently.
Generalized linear models, such as logistic regression, are a special form of GAMs where each $f_i$ is restricted to be linear.

\paragraph{Preliminaries: decision trees}
CART~\cite{breiman1984classification} fits a binary decision tree via recursive partitioning.
When growing a tree, CART chooses for each node $\node$ the split $s$ that maximizes the impurity decrease in the responses $\by$.
For a given node $t$, the impurity decrease has the expression 

\begin{equation}
\label{eq:tree}
\hat{\Delta}(s, \mathfrak{t}, \mathbf{y}):=  \sum_{\mathbf{x}_i \in \mathfrak{t}} h\left(y_i, \bar{y}_{\mathfrak{t}}\right)-\sum_{\mathbf{x}_i \in \mathfrak{t}_L} h\left(y_i,\bar{y}_{\mathfrak{t}_L}\right) -\sum_{\mathbf{x}_i \in \mathfrak{t}_R} h\left(y_i, \bar{y}_{\mathfrak{t}_R}\right),
\end{equation}

where $\node_L$ and $\node_R$ denote the left and right child nodes of $\node$ respectively, and $\bar y_{\node}, \bar y_{\node_L}, \bar y_{\node_R}$ denote the mean responses in each of the nodes.
For classification, $h(\cdot, \cdot)$ corresponds to the Gini impurity, and for regression, $h(\cdot, \cdot)$ is the mean-squared error.
Each split $s$ is a partition of the data based on a feature in $X$. 
To grow the tree, the splitting process is repeated recursively for each child node until a stopping criteria (e.g. a max depth) is satisfied.
At inference time, we predict the response of an example by following its path from the root to a leaf and then predicting with the mean value found in that leaf.

\subsection{\methods method description}
\label{subsec:gam_methods}

To remedy the issues with the GAM model in \cref{eq:GAM}, we propose \method, an intuitive model which leverages a pre-trained LLM to extract a feature representation $\phi(x_i)$ for each input ngram $x_i$.
This allows learning only a single linear weight vector $w$ with a fixed dimension (which depends on the embedding dimension produced by the LLM), regardless of the number of ngrams.
As a result, \methods can learn efficiently as the number of input features grows, and can also infer coefficients for unseen features.
The fitted model is still a GAM, ensuring that the model can be precisely interpreted as a linear function of its inputs:
\begin{equation}
    \label{eq:emb_gam}
    g(\mathbb{E}[y])=\beta+w^T \sum_i \phi(x_i)
\end{equation}
Fitting \methods is similar to the popular approach of finetuning a single linear layer on top of LLM embeddings.
However, it requires extra steps that separately extract/embed each ngram to keep the contributions to the prediction strictly additive across ngrams (see \cref{fig:intro}A):
\textit{(i) Extracting ngrams:}
To ensure input ngrams are interpretable, ngrams are constructed using a word-level tokenizer (here, spaCy~\cite{spacy2}).
We select the size of ngrams to be used via cross-validation.
\textit{(ii) Extracting embeddings:} 
Each ngram is fed through the LLM to retrieve a fixed-size embedding.\footnote{If a transformer returns a variable-length embedding (e.g. the embedding is the size of the sequence length), we average over its variable-length dimension. A common alternative for bi-directional (masked) language models is to use the embedding for a special token (i.e. \texttt{[CLS]}), but we aim to keep the approach here more general.}
\textit{(iii) Summing embeddings:}
The embeddings of each ngram in the input are summed to yield a single fixed-size vector, ensuring additivity of the final model.
\textit{(iv) Fitting the final linear model to make predictions:}
A linear model is fit on the summed embedding vector.
We choose the link function $g$ to be the logit function (or the softmax for multi-class) for classification and the identity function for regression.
In both cases, we add $\ell_2$ regularization over the parameters $w$ in \cref{eq:emb_gam}.

\paragraph{Computational considerations}
During fitting, \methods is inexpensive to fit as
(1) the pre-trained LLM is not modified in any way, and can be any existing off-the-shelf model
and 
(2) \methods only requires fitting a fixed-size linear model.
After training, the model can be converted to a dictionary of scalar coefficients for each ngram, 
where the coefficient is the dot product between the ngram's embedding and the fitted weight vector $w$ (\cref{fig:intro}B).
This makes inference extremely fast and converts the model to have size equal to the number of fitted ngrams.
When new ngrams are encountered at test-time, the coefficients for these ngrams can optionally be inferred by again taking the dot product between the ngram's embedding and the fitted weight vector $w$;

\subsection{\methodts method description}
\label{subsec:tree_methods}
\methodts improves upon a CART decision tree by augmenting features with generations from an LLM.
This helps capture correlations between ngrams,
including correlations with ngrams that are not present in the training data.
\methodts does not modify the objective in \cref{eq:tree} but rather modifies the procedure for fitting each individual split $s$ (\cref{fig:intro}D).
While CART restricts each split to a single ngram,
\methodts lets each split fit a \textbf{disjunction of ngrams} (e.g. \textit{ngram1 $\lor$ ngram2 $\lor$ ngram3}).
The disjunction allows a split to capture sparse interactions, such as synonyms in natural language.
This can help mitigate overfitting by allowing individual splits to capture concrete concepts,
rather than requiring many interacting splits.

When fitting each split, \methodts starts with the ngram which maximizes the objective in \cref{eq:tree}, just as CART would do, e.g. \textit{not good}.
Then, we query an LLM to generate similar ngrams to include in the split, e.g. \textit{bad, poor, awful, ..., horrendous}.
Specifically, we query GPT-3 (\texttt{text-davinci-003})~\cite{brown2020language} with the prompt \textit{Generate 100 concise phrases that are very similar to the keyphrase:\textbackslash nKeyphrase: ``\{keyphrase\}''\textbackslash n1.} and parse the outputs into a list of ngrams.
We greedily screen each ngram by evaluating the impurity of the split when including the ngram in the disjunction;
we then exclude any ngram which increases the split's impurity, resulting in a shortened list of ngrams, e.g. \textit{bad, poor, dull}.
See extended algorithm details in \cref{alg:method}.

\paragraph{Computational considerations}
As opposed to \method, \methodts uses an LLM API rather than LLM embeddings, which may be more desirable depending on access to compute.
The number of LLM calls required is proportional to the number of nodes in the tree.
During inference, the LLM is no longer needed and making a prediction simply requires checking an input for the presence of specific ngrams along one path in the tree.
Storing an \methods model requires memory proportional to the number of raw strings associated with tree splits, usually substantially reducing memory over the already small \methods model.
We explore variations of \methodts (such as using LLM embeddings rather than an API) in \cref{sec:app_tree}.

\section{Results: Prediction performance}
\label{sec:results}

\subsection{Experimental setup}
\label{subsec:experimental_setup}

\paragraph{Datasets}
\cref{tab:datasets_ovw} shows the datasets we study:
4 widely used text classification datasets spanning different domains (e.g. classifying the emotion of tweets~\cite{saravia2018carer},
the sentiment of financial news sentences~\cite{Malo2014GoodDO},
or the sentiment of movie reviews~\cite{PangLee2005,socher2013recursive}),
and 1 scientific text regression dataset (described in \cref{sec:fMRI})~\cite{lebel2022natural}.
Across datasets, the number of unique ngrams grows quickly from unigrams to bigrams to trigrams.
Moreover, many ngrams appear very rarely, e.g., in the Financial Phrasebank (FPB) dataset, 91\% of trigrams appear only once in the training dataset.

\begin{table}[t]
    \centering
    \scriptsize
    \caption{Overview of datasets studied here.
    The number of ngrams grows quickly with the size of the ngram.
    }
    \input{tabs/datasets_ovw}
    \label{tab:datasets_ovw}
\end{table}

\paragraph{\methods settings}
We compare \methods to four interpretable baseline models:
Bag of ngrams,
TF-IDF (Term frequency–inverse document frequency)~\cite{jones1972statistical},
GloVE~\cite{pennington2014glove}\footnote{We use the pre-trained Glove embeddings trained on Common Crawl (840 billion tokens, 2.2 million vocab-size, cased, 300-dimensional vectors).},
and a model trained on BERT embeddings for each unigram in the input (which can be viewed as running \methods with only unigrams).
We use BERT (\texttt{bert-base-uncased})~\cite{devlin2018bert} as the LLM for extracting embeddings, after finetuning on each dataset.\footnote{Pre-trained language models are retrieved from HuggingFace~\cite{wolf2019huggingface}. See \cref{tab:pretrained_models} for details on all models and downloadable checkpoints.}
In each case, a model is fit via cross-validation on the training set (to tune the amount of $\ell_2$ regularization added) and its accuracy is evaluated on the test set.

\paragraph{\methodts settings} We compare \methodts to two decision tree baselines: CART~\cite{breiman1984classification} and ID3~\cite{quinlan1986induction},
and we use bigram features.
In addition to individual trees, we fit bagging ensembles, where each tree is created using a bootstrap sample the same size as the original dataset (as done in Random Forest~\cite{breiman2001random}) and has depth 8.
This hurts interpretability, but can improve predictive performance and calibration.
For simplicity, we run \methods only in a binary classification setting; to do so, we take two opposite classes from each multiclass dataset (\textit{Negative}/\textit{Positive} for \textit{FPB} and \textit{Sadness}/\textit{Joy} for \textit{Emotion}).

\subsection{\methods text-classification performance}
\label{subsec:results_gam}
\begin{figure}[t]
    \centering
    \begin{tabular}{cc}
    \begin{overpic}[width=0.4\textwidth,tics=10]{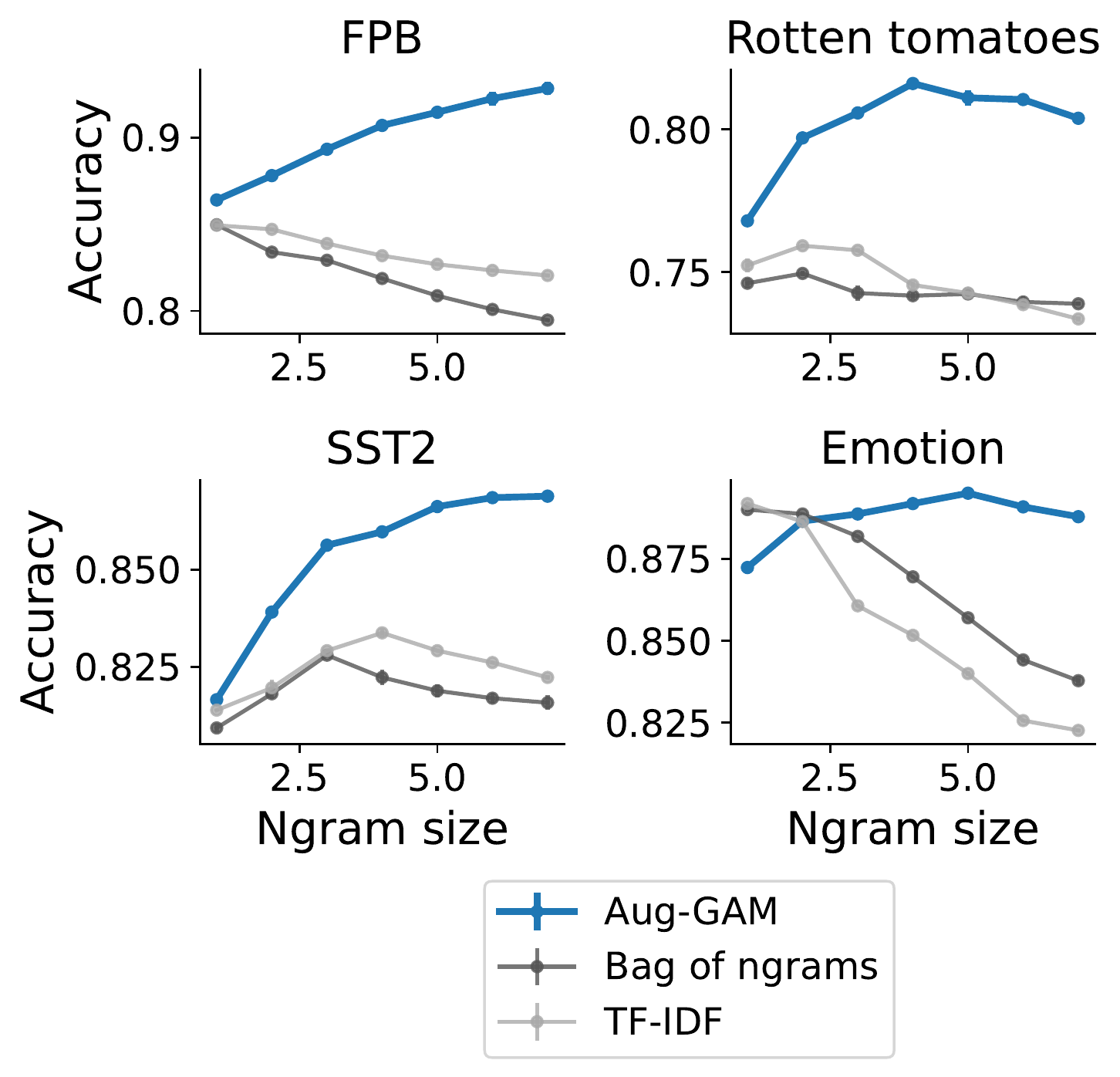}
        \put (2,95) {\large \textbf{A}}
    \end{overpic}
    & \begin{overpic}[width=0.45\textwidth,tics=10]{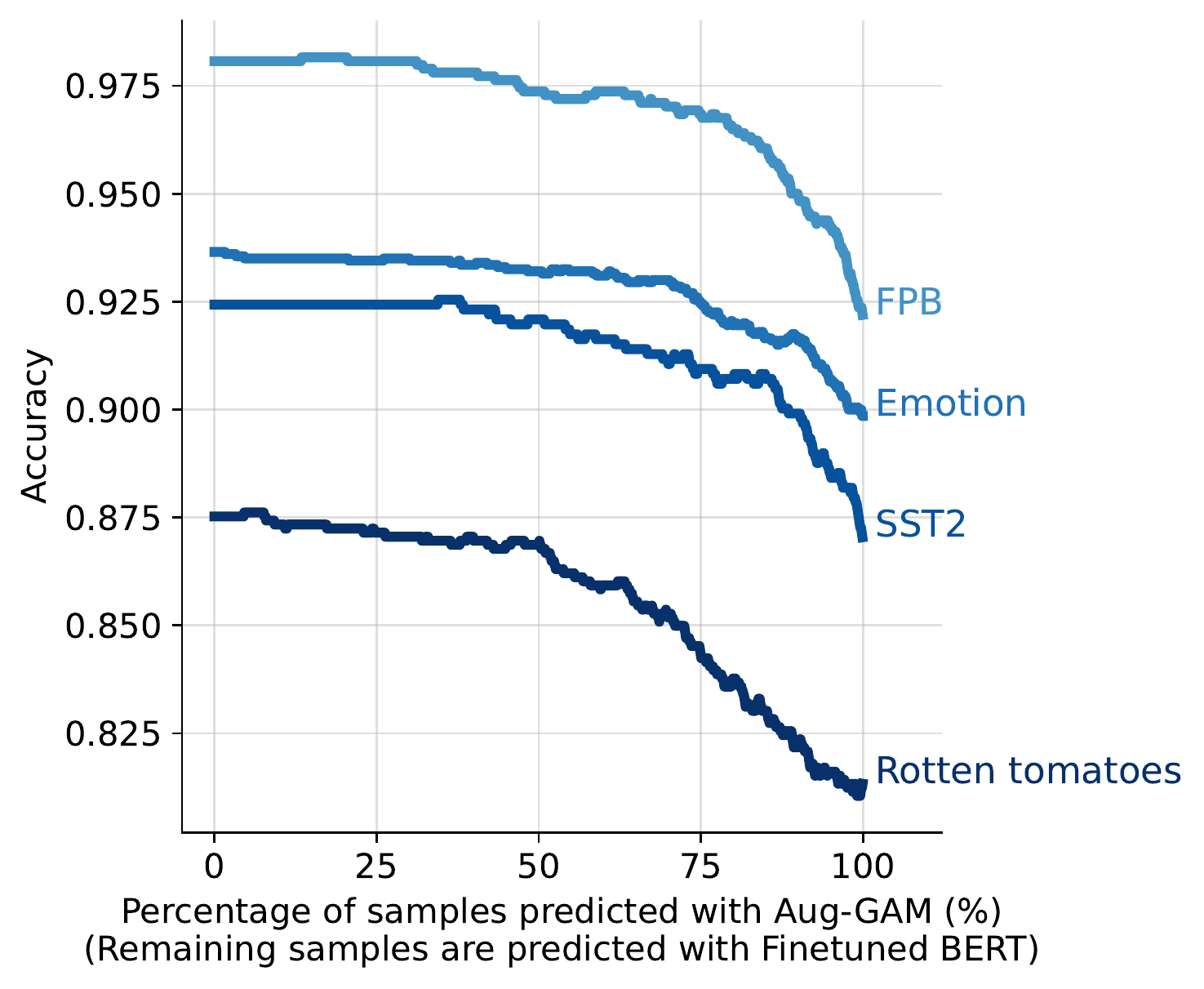}
        \put (0,83) {\large \textbf{B}}
    \end{overpic}
    \end{tabular}
\caption{
\textbf{(A)}
Test accuracy as a function of ngram size.
As the ngram size (i.e. the number of tokens in the ngram) increases, the gap between \methods and the baselines grows.
Averaged over three random cross-validation splits; error bars are standard errors of the mean (many are within the points).
\textbf{(B)} Accuracy when using \methods in combination with BERT.
A large percentage of samples can be accurately predicted with \method.
}
    \label{fig:generalization_vs_ngram_size}
\end{figure}

\paragraph{Generalization as a function of ngram size} \cref{fig:generalization_vs_ngram_size}A shows the test accuracy of \methods as a function of the ngram size used for computing features.
\methods outperforms the interpretable baselines, achieving a considerable increase in accuracy across three of the four datasets.
Notably, \methods accuracy increases with ngram size, whereas the accuracy of baseline methods decreases or plateaus.
This is likely due to \methods fitting only a fixed-size parameter vector, helping to prevent overfitting.
  
\paragraph{Comparing \methods performance with black-box baselines}
\cref{tab:best_results} shows the test accuracy results for various models when choosing the size of ngrams via cross-validation.
Compared with interpretable baselines, \methods shows considerable gains on three of the datasets
and only a minor gain on the tweet dataset (\textit{Emotion}), likely because this dataset requires fitting less high-order interactions.

Compared with the zero-shot performance of the much larger GPT models (6-billion parameter GPT-J~\cite{gpt_j} and 175-billion parameter GPT-3, \texttt{text-davinci-002}~\cite{brown2020language})\footnote{Accuracy for GPT models is computed by averaging over human-written prompts take from PromptSource (\cite{bach2022promptsource}); see details in \cref{sec:appendix}).},
\methods outperforms GPT-J.
\methods lags slightly behind GPT-3 for binary classification problems (\textit{Rotten Tomatoes} and \textit{SST2}) but outperforms GPT-3 for multi-class classification problems (\textit{FPB} and \textit{Emotion}).
The best black-box baseline (a BERT finetuned model) outperforms \methods by 4\%-6\% accuracy.
This is potentially a reasonable tradeoff in settings where interpretability, speed, or memory bottlenecks are critical.

\begin{table}[t]
    \centering
    \scriptsize
    \caption{
    Test accuracy for different models.
    \methods yields improvements over interpretable baselines and is competitive with some black-box baselines.
    Errors show standard error of the mean over 3 random data splits (or 3 different prompts for GPT models).
    }
    \input{tabs/best_results}
    \label{tab:best_results}
\end{table}

\paragraph{Complementing \methods with a black-box model}
In some settings, it may be useful to use \methods on relatively simple samples (for interpretability/memory/speed) but relegate relatively difficult samples to a black-box model.
To study this setting, we first predict each sample with \method, then assess its confidence (how close its predicted probability for the top class is to 1).
If the confidence is above a pre-specified threshold, we use the \methods prediction.
Otherwise, we compute the sample's prediction using a finetuned BERT model.
\cref{fig:generalization_vs_ngram_size}B shows the accuracy for the entire test set as we vary the percentage of samples predicted with \method.
Since \methods yields probabilities that are reasonably calibrated (see \cref{fig:acc_calibration}), rather than the accuracy linearly interpolating between \methods and BERT, a large percentage of samples can be predicted with \methods while incurring little to no drop in accuracy.
For example, when using \methods on 50\% of samples, the average drop in test accuracy is only 0.0053.

\paragraph{Tradeoffs between accuracy and efficiency}

In cases involving inference memory/speed, \methods can be converted to a dictionary of coefficients, whose size is the number of ngrams that appeared in training (see \cref{tab:datasets_ovw}).
For a trigram model, this yields roughly a 1,000x reduction in model size compared to the $\sim$110 million trainable parameters in BERT, with much room for further size reduction (e.g. simply removing coefficients for trigrams that appear only once yields another 10-fold size reduction).
Inference is nearly instantaneous, as it requires looking up coefficients in a dictionary and then a single sum (and does not require a GPU).

\cref{subsec:efficiency_tradeoffs_gam} explores accuracy/efficiency tradeoffs.
For example, \methods performance degrades gracefully when the model is compressed by removing its smallest coefficients.
In fact, the test accuracy of \methods models trained with 4-grams on the \textit{Emotion} and \textit{Financial phrasebank} datasets actually improves after removing 50\% of the original coefficients (\cref{fig:vary_ngrams_single}A).
Additionally, one can vary the size of ngrams used at test-time without a severe performance drop, potentially enabling compressing the model by orders of magnitude (see \cref{fig:vary_ngrams_single}B, \cref{fig:vary_ngrams_test}).
For example, when fitting a model with 4-grams and testing with 3-grams, the average performance drop is $\sim$2\%.

\subsection{\methodts generalization performance}

We now investigate the predictive performance of \methodt,
measured by the test ROC AUC on the previous text classification datasets altered for binary classification.
Note that the performance of all tree-based methods on the studied datasets is below the performance of the GAM methods in \cref{subsec:results_gam} (see \cref{tab:variations_tree_acc} for a direct comparison).
Nevertheless, \methodts models maintain potential advantages,
such as storing far fewer parameters,
clustering important features together,
and better modeling long-range interactions.

\begin{figure}[t]
    \centering
    \begin{tabular}{ccl}
         \includegraphics[width=0.4\textwidth]{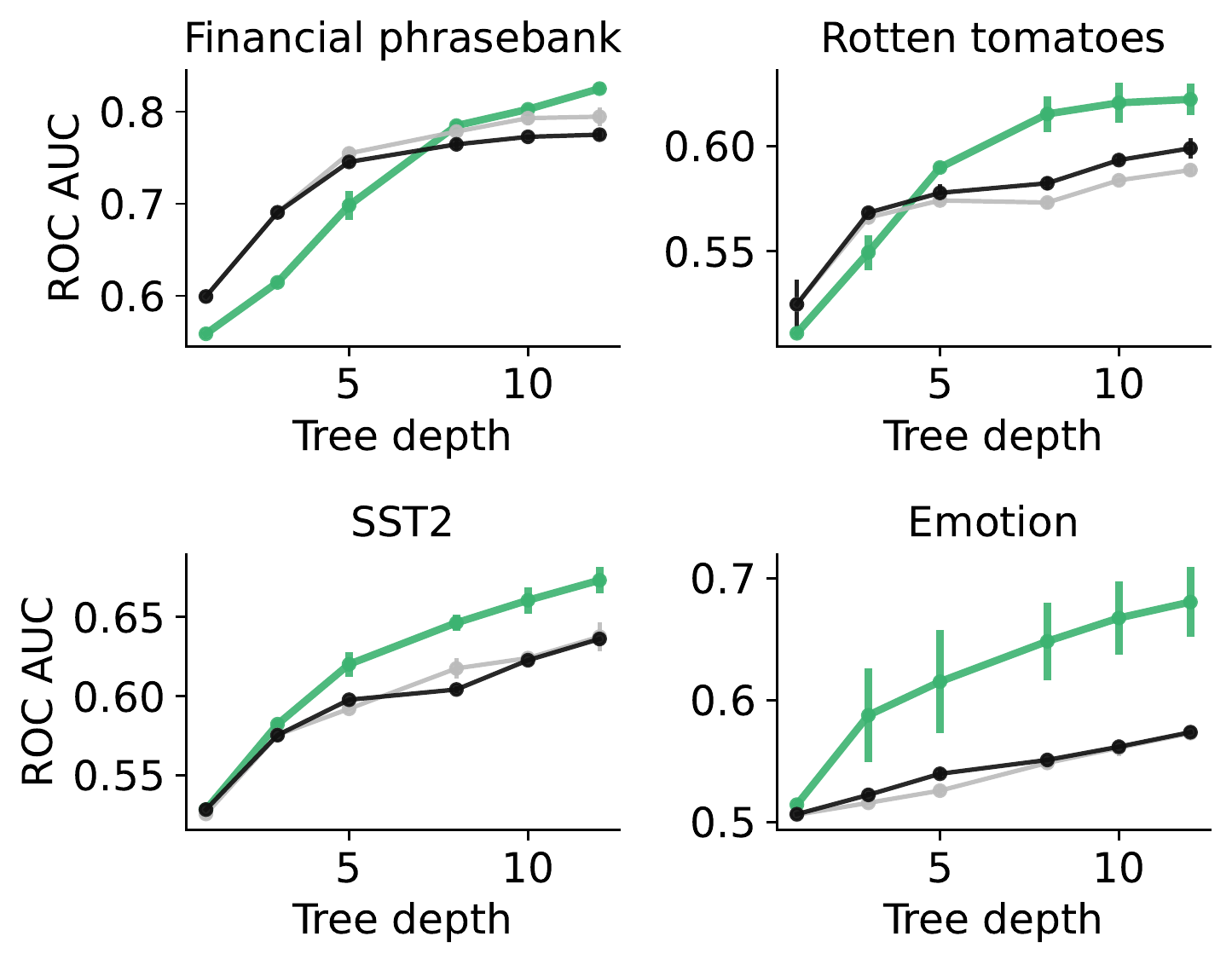} & 
         \includegraphics[width=0.4\textwidth]{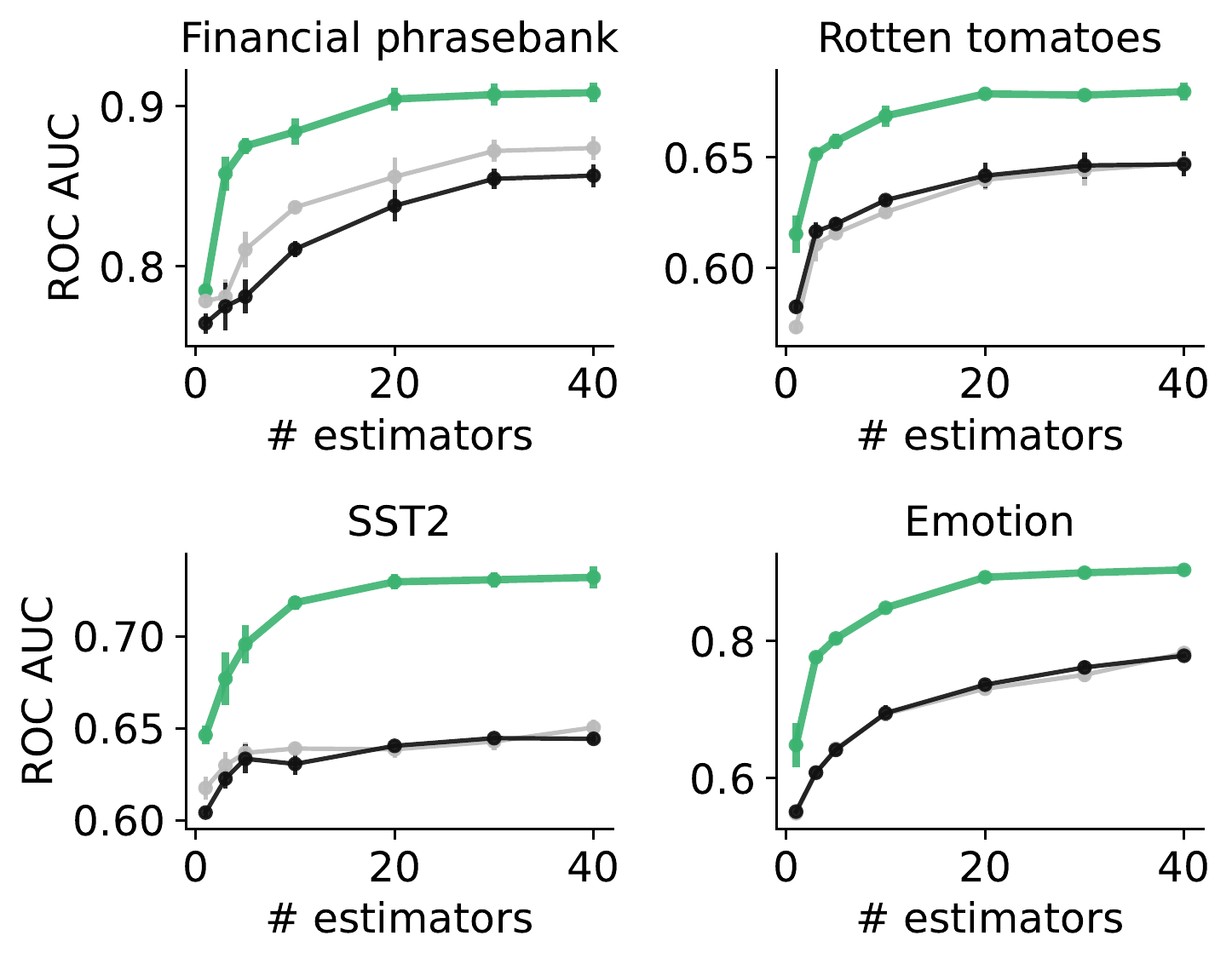} & 
         \makecell[t]{\includegraphics[width=0.09\textwidth]{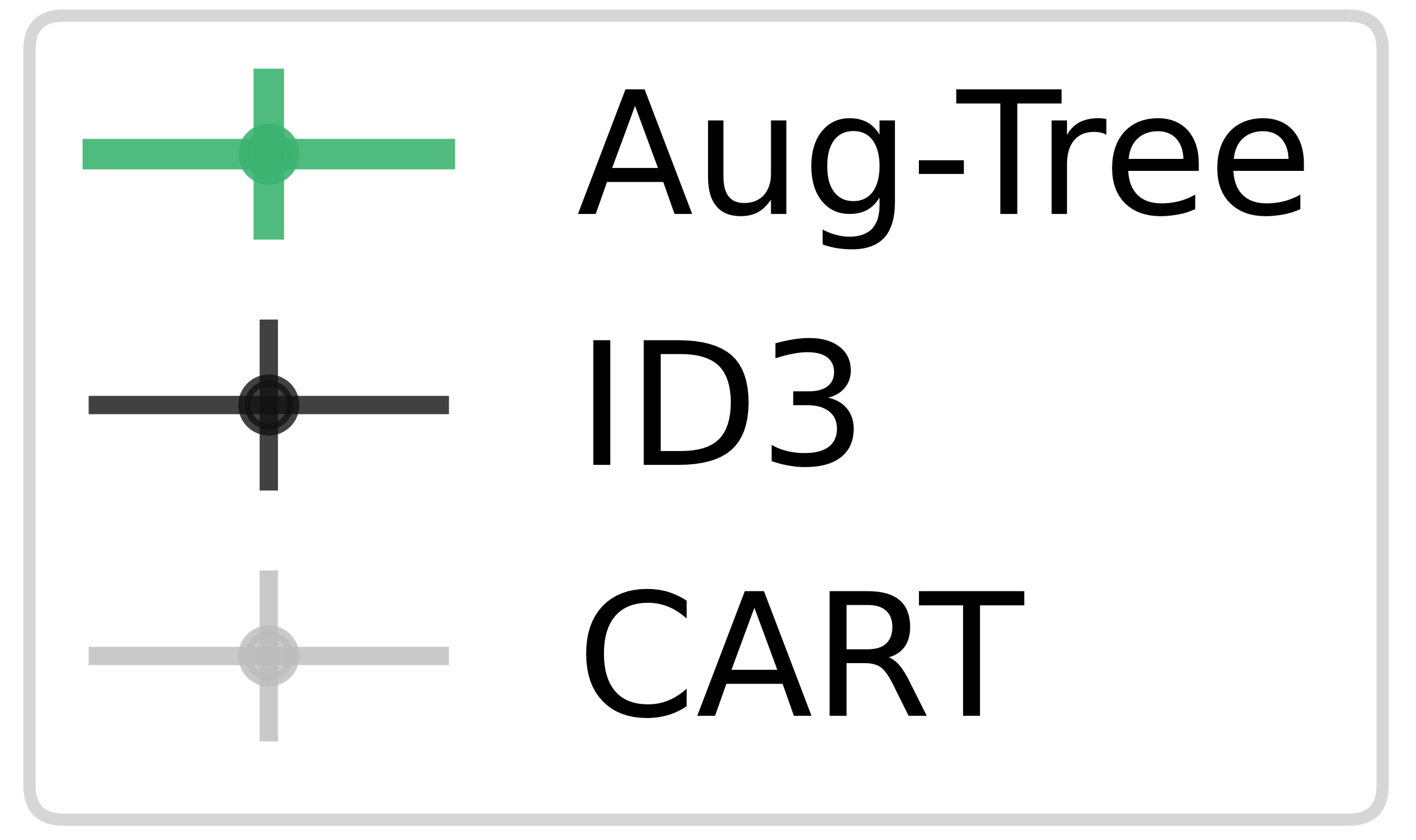}}\\
         \scriptsize{(A) Individual tree} & \scriptsize{(B) Bagging ensemble}
    \end{tabular}
    \caption{Test performance as a function of (A) tree depth and (B) number of estimators. Values are averaged over 3 random dataset splits; error bars show the standard error of the mean (many are within the points).}
    \label{fig:perf_curves_tree}
\end{figure}

\cref{fig:perf_curves_tree}A shows the performance for \methodts as a function of tree depth compared to decision tree baselines.
\methodts shows improvements that are sometimes small (e.g. for \textit{Financial phrasebank}) and sometimes relatively large (e.g. for \textit{Emotion}).
\cref{fig:perf_curves_tree}B shows the performance of a bagging ensemble of trees with different tree methods used as the base estimator.
Here, using \methodts shows a reliable and significant gain across all datasets compared to ensembles of baseline decision-tree methods.
This suggests that LLM augmentation may help to diversify or decorrelate individual trees in the ensemble.

\paragraph{Varying \methodmains settings}
We investigate many variations in the settings used for \methodmain.
\cref{tab:variations_tree} shows variations of different hyperparameters for \methodt, such as using embeddings or dataset-specific prompts to expand keyphrases.
\cref{tab:variation_results_full} shows how generalization accuracy changes when the LLM used to extract embeddings for \methods is varied, or different layers / ngram selection techniques are used.
Across the variations, embeddings from finetuned models yield considerably better results than embeddings from non-finetuned models.

\section{Interpreting fitted models}
\label{sec:interpretation}
In this section, we interpret fitted \methodmain.
We first inspect an \methods model fitted using unigram and bigram features
on the \textit{SST2} dataset which achieves 84\% test accuracy.
Next, we analyze the keyphrase expansions made in fitted \methodts bagging ensembles.

\begin{figure}[t]
    \centering
    \begin{tabular}{cc}
    \begin{overpic}[width=0.66\textwidth,tics=10]{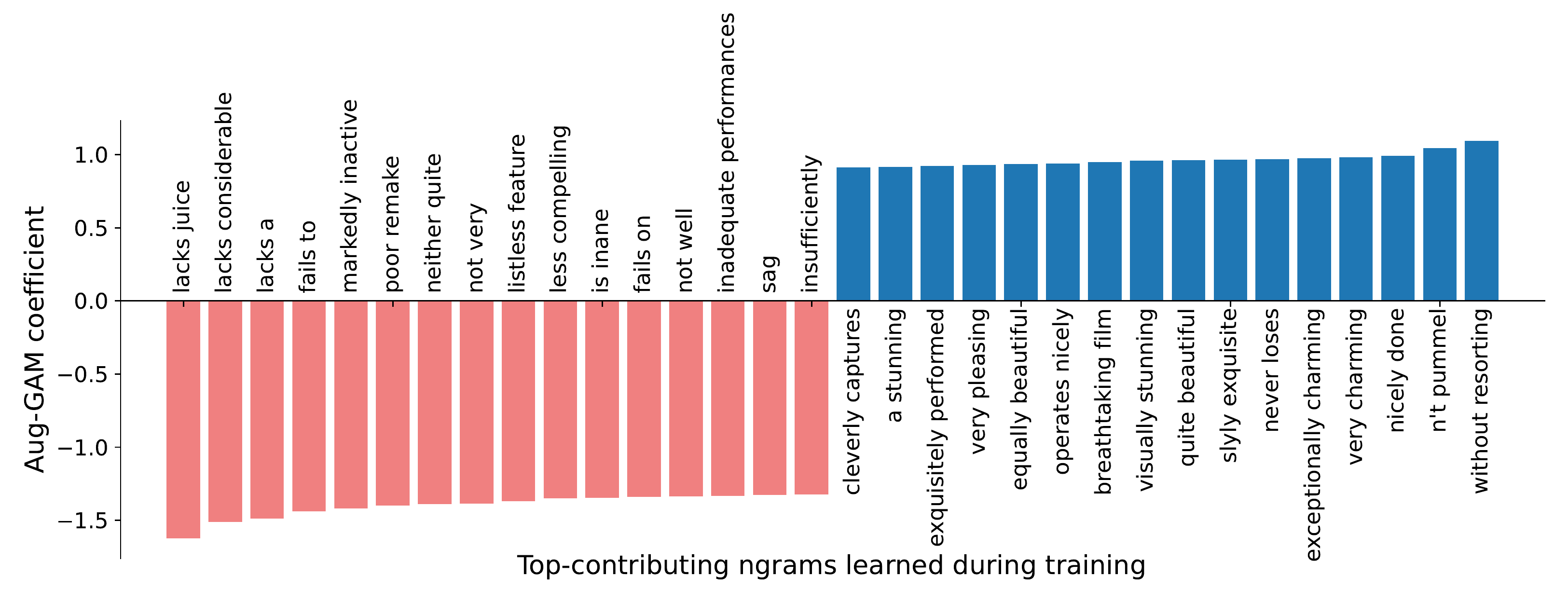}
        \put (1,30) {\large \textbf{A}}
    \end{overpic} & 
    \hspace{-15pt}
    \begin{overpic}[width=0.33\textwidth,tics=10]{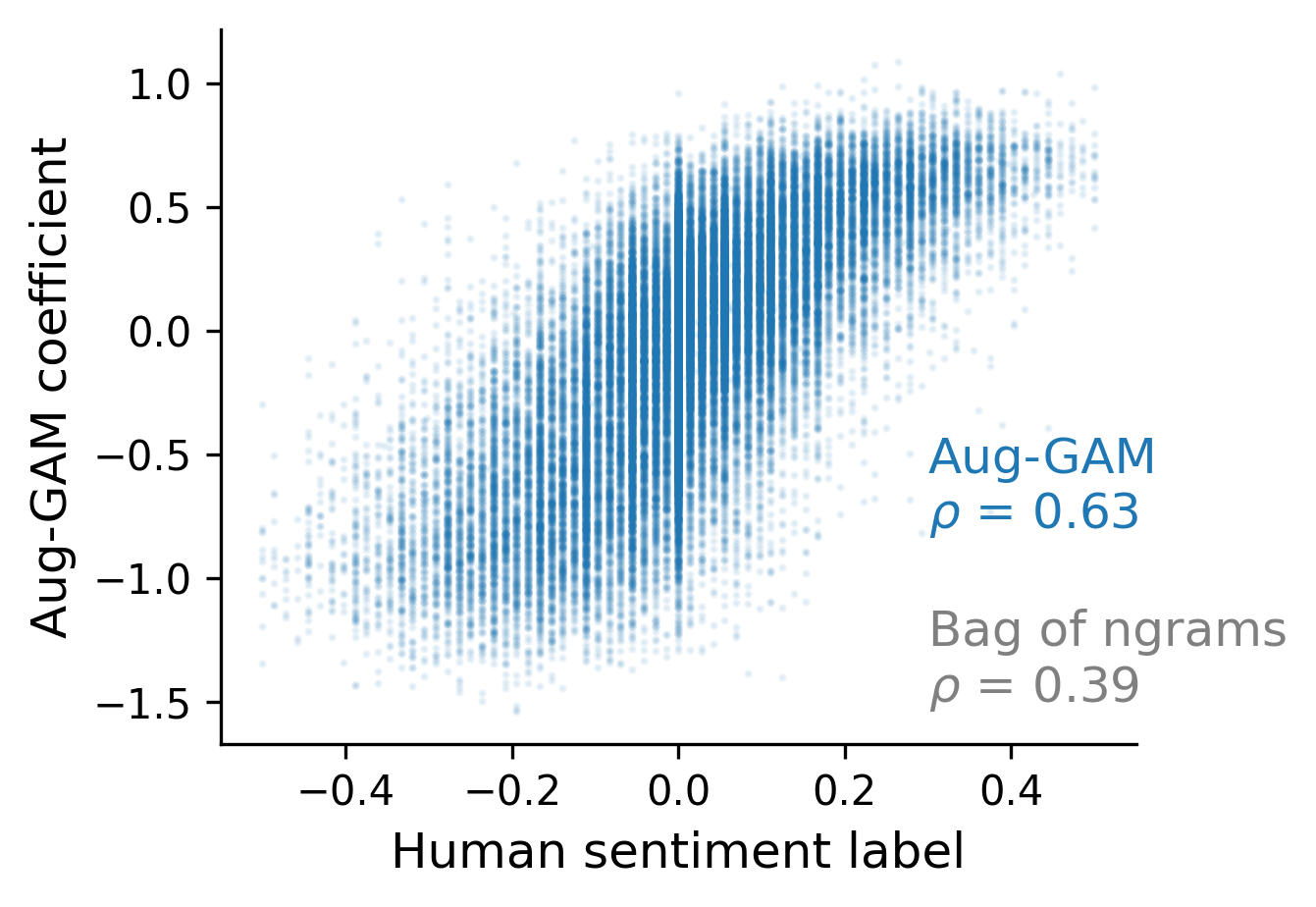}
        \put (1,66) {\large \textbf{B}}
    \end{overpic}\\
    \begin{overpic}[width=0.66\textwidth,tics=10]{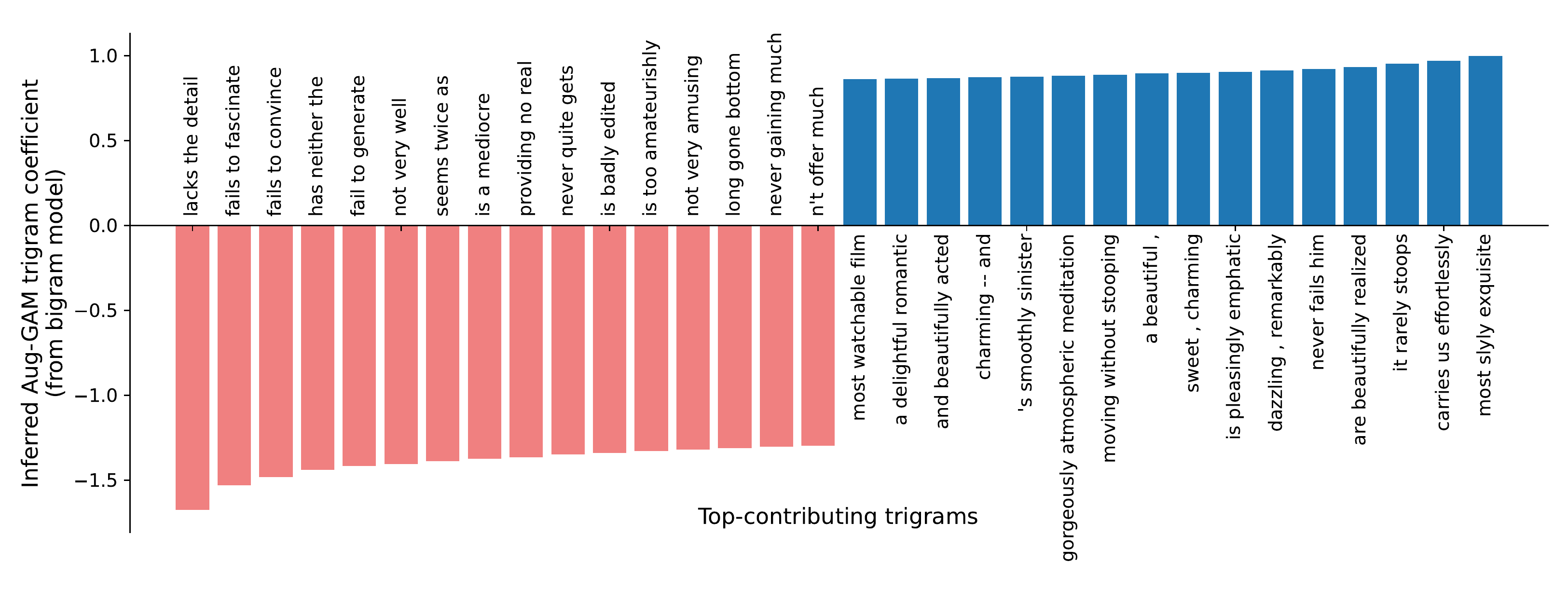}
        \put (1,35) {\large \textbf{C}}
    \end{overpic} & 
    \hspace{-15pt}
    \begin{overpic}[width=0.33\textwidth,tics=10]{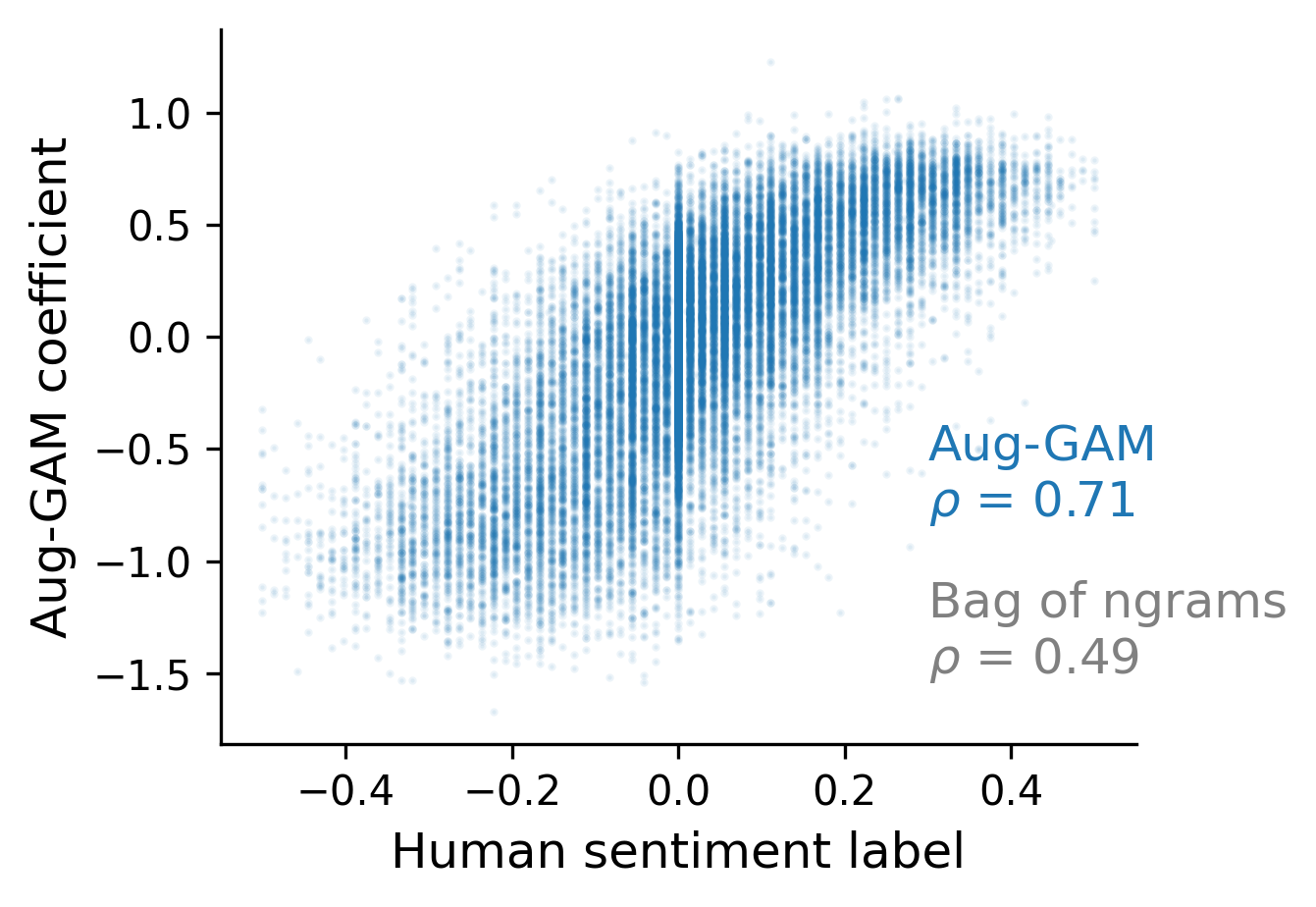}
        \put (1,66) {\large \textbf{D}}
    \end{overpic}
    \end{tabular}
    \caption{Top and bottom contributing ngrams to an \methods model trained on SST2 bigrams are \textbf{(A)} qualitatively semantically accurate and \textbf{(B)} match human-labeled phrase sentiment scores.
    For the same \methods model, which is trained only on bigrams, 
    inferred trigrams coefficients are
    \textbf{(C)} qualitatively semantically accurate and
    \textbf{(D)} match human-labeled phrase sentiment scores.}
    \label{fig:ngram_contribution}
\end{figure}

\paragraph{Fitted \methods coefficients match human scores}
A fitted \methods model can be interpreted for a single prediction (i.e. getting a score for each ngram in a single input, as in \cref{fig:intro}) or for an entire dataset (i.e. by inspecting its fitted coefficients).
\cref{fig:ngram_contribution}A visualizes the fitted \methods coefficients (i.e. the contribution to the prediction $w^T \phi(x_i)$) with the largest absolute values across the SST2 dataset.
To show a diversity of ngrams, we show every fifth ngram.
The fitted coefficients are semantically reasonable and many contain strong interactions (e.g. \textit{not very} is assigned to be negative whereas \textit{without resorting} is assigned to be positive).
This form of model visualization allows a user to audit the model with prior knowledge.
Moreover, these coefficients are exact and therefore avoid summarizing interactions, making them considerably more faithful than post-hoc methods, such as LIME~\cite{ribeiro2016should} and SHAP~\cite{lundberg2016unexpected} (see \cref{subsec:posthoc_comparison} for a comparison).

\cref{fig:ngram_contribution}B compares the fitted \methods coefficients to human-labeled sentiment phrase scores for unigrams/bigrams in SST (note: these continuous scores are separate from the binary sentence labels used for training in the SST2 dataset).
Both are centered, so that 0 is neutral sentiment and positive/negative values correspond to positive/negative sentiment, respectively.
There is a strong positive correlation between the coefficients and the human-labeled scores (Spearman rank correlation $\rho=0.63$), which considerably improves over coefficients from a bag-of-bigrams model trained on SST2 ($\rho=0.39$).

\paragraph{Inferred \methods coefficients for unseen ngrams match human scores}

One strength of \methods is its ability to infer linear coefficients for ngrams that were not seen during training.
Whereas baseline models generally assign each unknown ngram the same coefficient (e.g. 0),
\methods can effectively assign these new ngrams accurate coefficients.
As one example, \cref{fig:ngram_contribution}C shows that the \methods model trained only on bigrams in \cref{fig:ngram_contribution}A/B can automatically infer coefficients for trigrams (which were not fit during training).
The inferred coefficients are semantically meaningful, even capturing three-way interactions, such as \textit{not very amusing}.
To show a diversity of ngrams, we show every 20th ngram.
\cref{fig:ngram_contribution}D shows the coefficients compared to the human-labeled SST phrase sentiment for all trigrams in SST.
Again, there is a strong correlation, where the \methods coefficients achieves a rank correlation $\rho=0.71$, which even outperforms the bag-of-words model directly trained on trigrams ($\rho=0.49$).

\paragraph{\methodts augmented splits contain relevant phrases}
\label{subsec:llm_phrases}

A fitted \methodts model can be easily interpreted for a single prediction (i.e. by inspecting the ngrams that triggered relevant splits)
or by visualizing the entire tree (e.g. \cref{fig:intro}C).
Here, we additionally analyze how well each ngram found by CART matches the augmented ngrams found by the LLM;
the better this match is, the easier it is to interpret a split.

\cref{tab:expand_examples} shows examples of the ngrams which were most frequently augmented when fitting a bagging ensemble of 40 \methodt s to the four text-classification datasets in \cref{tab:datasets_ovw}.
Added ngrams seem qualitatively reasonable, e.g. the keyphrase \textit{good} expands to \textit{fine, highly, solid, ..., valuable}.
We evaluate how well the expansions match the original CART ngram via human evaluation scores.
Human evaluators are given the original ngram and the added ngrams, then instructed ``You are given a keyphrase along with related keyphrases. On a scale of 1 (worst) to 5 (best), how well do the related keyphrases match the example keyphrase?''\footnote{Human evaluation scores are averaged over 3 PhD students in machine learning not affiliated with the study and 15 random ngrams from each dataset.}.
\cref{tab:expand_examples} shows that the average human score for splits in each dataset is consistently greater than 4.
This is substantially higher than the baseline score of 1.3 assigned by human evaluators when 15 ngrams and expansions are randomly paired and evaluated.
\cref{tab:expansion_metadata} gives more details on ngram expansions.

\begin{table}[t]
    \centering
    \footnotesize
    \caption{Examples of most frequently augmented ngrams for each dataset when fitting an ensemble of 40 \methodt s.
    Human scores measure the similarity between an ngram and its expansion.
    They range from 1 (worst match) to 5 (best match), and 
    the baseline score when ngrams and expansions are randomly paired and evaluated is 1.3{\err{0.1}}.
    Error bars show standard error of the mean.
    Abbreviations: FPB = Financial Phrasebank, RT = Rotten tomatoes.
    }
    \def\arraystretch{1.2}
    \input{tabs/expansion_examples}
    \label{tab:expand_examples}
\end{table}

\section{Analyzing fMRI data with \methodmain}
\label{sec:fMRI}
We now explore \methodmains in a real-world neuroscience context.
A central challenge in neuroscience is understanding how and where semantic concepts are represented in the brain.
To meet this challenge, one line of study predicts the response of different brain voxels (i.e. small regions in space) to natural-language stimuli.
We analyze data from a recent study in which the authors collect functional MRI (fMRI) responses as human subjects listen to hours of narrative stories~\cite{lebel2022natural}.
The fMRI responses studied here contain 95,556 voxels from a single subject,
with 9,461 time points used for training/cross-validation and 291 time points used for testing.
We predict the continuous response for each voxel at each time point using the 20 words that precede the time point.\footnote{The most recent 4 words are skipped due to a time delay in the fMRI BOLD response.}
Seminal work on this task found that linear models of word vectors could effectively predict voxel responses~\cite{huth2016natural}, and more recent work shows that LLMs can further improve predictive performance~\cite{schrimpf2021neural,antonello2022predictive}.
\methods is well-suited to this task, as it combines low-level word information with the contextualized information present in higher-order ngrams, both of which have been found to contribute to fMRI representations of text~\cite{caucheteux2022brains}.

\begin{figure}[t]
    \centering
    \footnotesize
    \includegraphics[width=\textwidth]{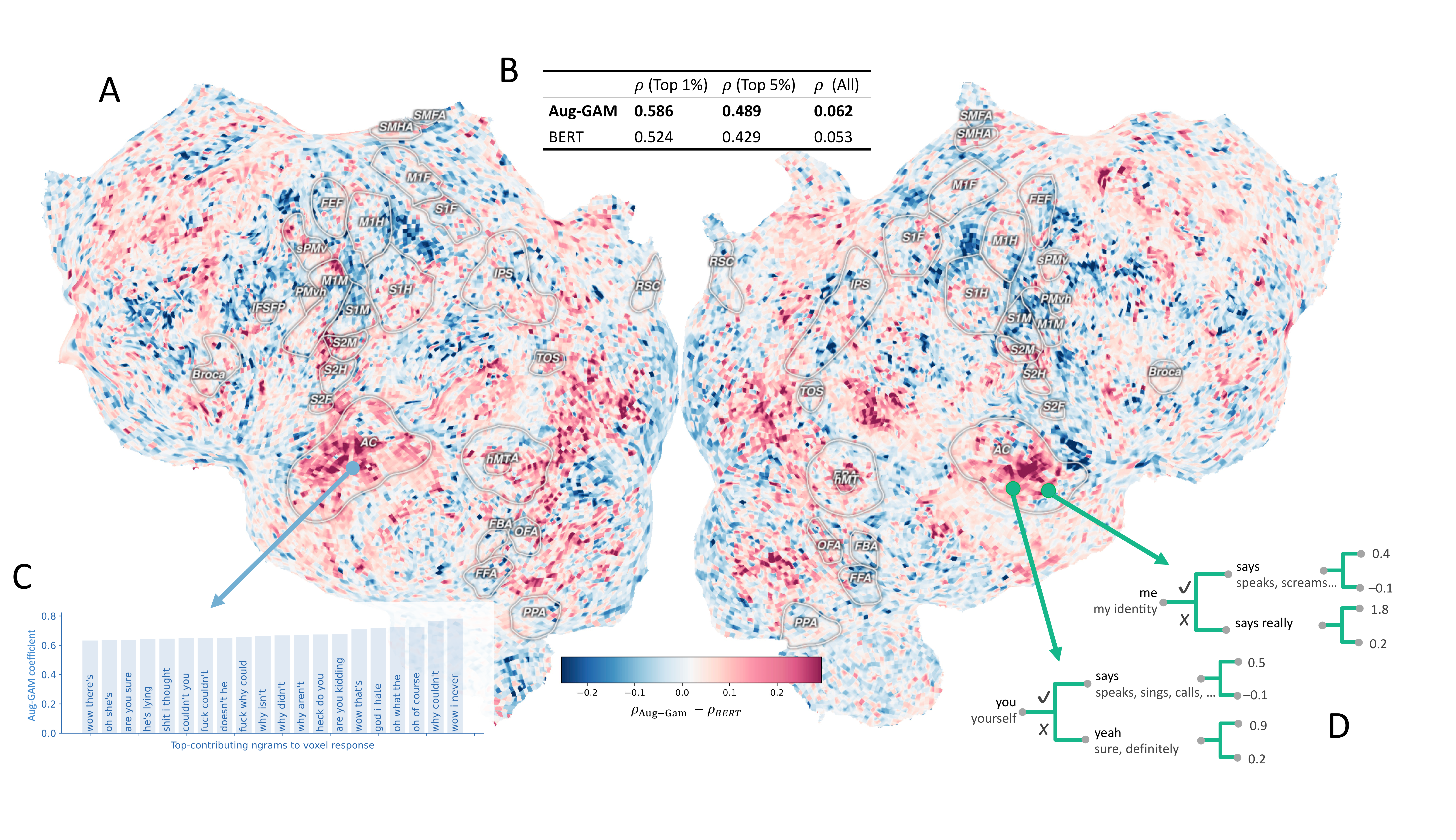}
    \vspace{-24pt}
    \caption{
    \methodmains prediction performance and interpretation for fMRI voxels.
    \textbf{(A)}
    Map of the difference between the performance of \methods and BERT for fMRI voxel prediction across the cortex.
    Positive values (red) show where \methods outperforms BERT (measured by correlation on the test set).
    \textbf{(B)}
    \methods outperforms BERT when averaging across all voxels (or just over the 1\%/5\% with the highest test correlations).
    Standard errors of the mean are all less than 0.0015.
    \textbf{(C)} Example \methods model for a single voxel (visualized with the top \methods coefficients).
    \textbf{(D)} Example \methodts model for two voxels.
    }
    \label{fig:flatmap_diff}
\end{figure}

\cref{fig:flatmap_diff}A visualizes the voxels in the cortex which are better predicted by \methods than BERT.
The improvements are often spatially localized within well-studied brain regions such as auditory cortex (AC).
\cref{fig:flatmap_diff}B shows that the test performance for \methods (measured by the Pearson correlation coefficient $\rho$) outperforms the black-box BERT baseline.
\cref{sec:fmri_supp} gives further data details and comparisons, e.g. \methods also outperforms other linear baselines.

Going beyond prediction performance, \cref{fig:flatmap_diff}C investigates a simple example of how \methods could help interpret an underlying brain region.
We first select the voxel which is best-predicted by \methods (achieving a test correlation of 0.76) and then visualize the largest fitted \methods coefficients for that voxel.
These correspond to which ngrams increase the activity of the fMRI voxel the most.
Interestingly, these ngrams qualitatively correspond to understandable concepts: \textit{questioning}, e.g. ``are you sure'', often combined with \textit{disbelief/incredulity}, e.g. ``wow I never''.
\cref{fig:flatmap_diff}D shows two examples of voxels that are better predicted by \methodts than \methods (\methodts yields test correlations of 0.35 and 0.36).
These two voxels are both related to someone speaking,
but they seem to depend on interactions between the noun (\textit{me} or \textit{you}) and the verb (\textit{says}).
To elicit a large response both must be present, something which is difficult to capture in additive models, even with ngrams, since these words may not be close together in a sentence.

This interpretation approach could be applied more rigorously to generate hypotheses for text inputs that activate brain regions,
and then testing them with followup fMRI experiments.

\FloatBarrier

\section{Background and related work}
\label{sec:background}

\paragraph{GAMs}
There is a large literature on additive models being used for interpretable modeling.
This includes generalized additive models (GAMs)~\cite{hastie1986generalized}, which have achieved strong performance in various domains by modeling individual component functions/interactions using regularized boosted decision trees~\cite{caruana2015intelligible} and more recently using neural networks~\cite{agarwal2021neural}.
However, existing GAM methods are limited in their ability to model the high-order feature interactions that arise in NLP.
Meanwhile, NLP has seen great success in models which build strong word-level representations, e.g. word2vec~\cite{mikolov2013efficient,mikolov2013distributed},
GloVe~\cite{pennington2014glove},
FastText~\cite{joulin2016bag}
and ELMo~\cite{peters-etal-2018-deep}.
By replacing such models with LLM embeddings, \methods enables easily modeling ngrams of different lengths without training a new model.
Moreover, unlike earlier methods, LLMs can incorporate information about labels into the embeddings (e.g. by first finetuning an LLM on a downstream prediction task).

\paragraph{Decision trees}
There is a long history of greedy methods for fitting decision trees, e.g., CART~\cite{breiman1984classification}, ID3~\cite{quinlan1986induction},
and C4.5~\cite{quinlan2014c4}.
More recent work has explored fitting trees via global optimization rather than greedy algorithms~\cite{lin2020generalized,hu2019optimal,bertsimas2017optimal}; this can improve performance given a fixed tree size but incurs a high computational cost.
Other recent studies have improved trees after fitting through regularization~\cite{agarwal2022Hierarchical}
or iterative updates~\cite{carreira2018alternating}.
Beyond trees, there are many popular classes of rule-based models, such as
rule sets~\cite{friedman2008predictive},
rule lists~\cite{angelino2017learning,singh2021imodels},
and tree sums~\cite{tan2022Fast}.
\methodts addresses a common problem shared by rule-based approaches: modeling the sparse, correlated features that are common in tasks such as text classification.

Beyond fitting a single tree, tree ensembles such as Random Forest~\cite{breiman2001random}, gradient-boosted trees~\cite{freund1996experiments},
XGBoost~\cite{chen2016xgboost},
and BART~\cite{chipman2010bart},
have all shown strong predictive performance in diverse settings.
These ensembling approaches are compatible with \methodt, as it can be used as the base estimator in any of these approaches.

\paragraph{Interpreting/distilling neural networks}

The work here is related to studies that aim to make neural networks more interpretable.
For example, models can make predictions by comparing inputs to prototypes~\cite{li2018deep,chen2019looks},
by predicting intermediate interpretable concepts~\cite{koh2020concept,yang2022language,ghosh2023route},
using LLMs to extract prompt-based features~\cite{yuksekgonul2022post,mcinerney2023chill},
distilling a neural network into a mostly transparent model~\cite{frosst2017distilling}
or distilling into a fully transparent model (e.g. adaptive wavelets~\cite{ha2021adaptive} or an additive model~\cite{tan2018learning}).
Separately, many works use neural network distillation to build more efficient (but still black-box) neural network models, e.g.~\cite{hinton2015distilling,sanh2019distilbert}.

\paragraph{Feature and feature-interaction importances}
Loosely related to this work are post-hoc methods that aim to help understand a black-box model,
i.e. by providing feature importances using methods such as LIME~\cite{ribeiro2016should}, SHAP~\cite{lundberg2019explainable}, and others \cite{friedman2001greedy,devlin2019disentangled}.
However, these methods lose some information by summarizing the model and suffer from issues with summarizing interactions~\cite{rudin2018please,murdoch2019Definitions}.
Slightly more related are works which aim to explain feature interactions or transformations in neural networks~\cite{janizek2021explaining,singh2019Hierarchical,singh2020transformation}, but these works fail to explain the model as a whole and are again less reliable than having a fully transparent model.

\section{Discussion}
\label{sec:discussion}

\methodmains provide a promising direction towards future methods that reap the benefits of both LLMs and transparent models in NLP: high accuracy along with interpretability/efficiency.
This potentially opens the door for introducing LLM-augmented models in high-stakes domains, such as medical decision-making and in new applications on compute-limited hardware.
\methodmains is currently limited to applications for which an effective LLM is available, and thus may not work well for very esoteric NLP tasks.
However, as LLMs improve, the predictive performance of \methodmains should continue to improve and expand to more diverse NLP tasks.
More generally, \methodmains can be applied to domains outside of NLP where effective foundation models are available (e.g. computer vision or protein engineering).

\methodmains can be readily extended to new model forms beyond additive models and trees.
Other transparent models, such as rule lists, rule sets, and prototype-based models could all potentially benefit from LLM augmentation during training time.
In all these cases, LLM augmentation could use LLM embeddings (as is done in \method),
use LLM generations (as is done in \methodt),
or use LLMs in new ways.
\methods could be augmented by building on the nonlinearity present in GAMs such as the explainable boosting machine~\cite{caruana2015intelligible},
to nonlinearly transform the embedding for each ngram with a model before summing to obtain the final prediction.
Additionally, \methods could fit long-range interaction terms as opposed to only ngrams.
\methodts could leverage domain knowledge to engineer more meaningful prompts for expanding ngrams or for extracting relevant ngrams.
Both models can be further studied to improve their compression (potentially with LLM-guided compression techniques)
or to extend their capabilities to tasks beyond classification/regression, such as sequence prediction or outlier detection.
We hope that the introduction of \methodmains can help push improved performance prediction into high-stakes applications, improve interpretability for scientific data, and reduce unnecessary energy/compute usage.


\FloatBarrier
{
    \input{_main.bbl}

}

\newpage
\begin{appendices}

\input{appendix_gam.tex}
\input{appendix_tree.tex}
\input{appendix_fmri}

\end{appendices}

\end{document}

%% file: preamble.tex
\usepackage{graphicx}
\usepackage{float}

\definecolor{mygray}{gray}{0.5}
\definecolor{cblue}{RGB}{8, 85, 153}
\definecolor{darkblue}{RGB}{1, 43, 112}

\definecolor{cgreen}{RGB}{8, 153, 83}

\usepackage{amsmath}
\usepackage{amssymb}
\usepackage{amsthm}
\usepackage{booktabs}
\usepackage{longtable}
\usepackage{placeins} 
\usepackage{makecell}
\usepackage{multirow}
\usepackage[percent]{overpic}

\usepackage[margin=1.0in]{geometry}

\usepackage[capitalize]{cleveref}
\Crefname{section}{Sec}{Secs.}
\Crefname{figure}{Fig}{Figs.}

\usepackage{fontawesome}

\newcommand{\methodmainlongs}{Augmented Interpretable Models }
\newcommand{\methodmain}{Aug-imodels}
\newcommand{\methodmainsingular}{Aug-imodel}

\newcommand{\methodmains}{Aug-imodels }
\newcommand{\method}{Aug-GAM}
\newcommand{\methods}{Aug-GAM }
\newcommand{\methodt}{Aug-Tree}
\newcommand{\methodts}{Aug-Tree }
\usepackage{adjustbox}

\usepackage{tabularx}
\newcommand{\tin}[1]{\tiny{\texttt{#1}}}

\definecolor{cmaroon}{RGB}{128, 0, 0}
\usepackage{algorithm}  
\usepackage{algorithmicx}
\usepackage[noend]{algpseudocode}
\newcommand{\var}[1]{\textcolor{darkgray}{\textit{#1}}}

\algdef{SE}[DOWHILE]{Do}{doWhile}{\algorithmicdo}[1]{\algorithmicwhile\ #1}%
\renewcommand\algorithmicdo{}
\newcommand{\CommentMain}[1]{\textcolor{mygray}{\# #1}}

\newcommand{\node}{\mathfrak{t}}

\newcommand{\by}{\mathbf{y}}

\newcommand{\err}[1]{\footnotesize $\pm${#1}}


%% file: tabs/datasets_ovw.tex
\begin{tabular}{lrrrr|r}
\toprule
{} & \makecell{FPB} & \makecell{Rotten\\tomatoes} &    SST2 & Emotion & fMRI\\ 
\midrule
Samples (train)         &                 2,313 &            8,530 &   67,349 &   16,000 & 9,461\\
Samples (val)           &                 1,140 &            1,066 &     872 &    2,000 & 291\\
Classes                 &                    3 &               2 &       2 &       6 & Regression\\
Unigrams                &                 7,169 &           16,631 &   13,887 &   15,165 & 4,980\\
Bigrams                 &                28,481 &           93,921 &   72,501 &  106,201 & 27,247\\
\vspace{-5pt}
Trigrams                &                39,597 &          147,426 &  108,800 &  201,404 & 46,834 \\
\makecell[lt]{Trigrams that\\appear only once} & \makecell[r]{\\91\%\\} & \makecell[r]{\\93\%\\} & \makecell[r]{\\13\%\\} & \makecell[r]{\\89\%\\} & \makecell[r]{\\71\%\\} \\
\bottomrule
\end{tabular}

%% file: tabs/best_results.tex
\begin{tabular}{llllll}
\toprule
& {} & FPB &      Rotten tomatoes &                 SST2 &              Emotion \\
\midrule
\multirow{1}{*}{Ours}
& \textbf{Aug-GAM}                        &  \textbf{92.8} \err{ 0.37} &  \textbf{81.6} \err{0.05} &   \textbf{86.9} \err{ 0.10} &  \textbf{89.5} \err{ 0.03} \\
\midrule
\multirow{5}{*}{\makecell[l]{Interpretable\\baselines}} & Bag of ngrams                          &  85.0 \err{ 0.11} &  75.0 \err{ 0.09} &   82.8 \err{ 0.00} &  89.0 \err{ 0.09} \\
& TF-IDF                                 &  84.9 \err{ 0.16} &  75.9 \err{ 0.06} &  83.4 \err{ 0.11}  &  89.2 \err{ 0.04} \\
& GloVe & 80.5 \err{ 0.06}	 & 78.7 \err{ 0.03}	& 80.1 \err{ 0.10} & 73.1 \err{ 0.09}\\
& \makecell[l]{BERT unigram embeddings}  &  86.4 \err{ 0.13} &  76.8 \err{ 0.19} &  81.7 \err{ 0.07} &  87.2 \err{ 0.06} \\
\midrule
\multirow{3}{*}{\makecell[l]{Black-box\\baselines}} & \textbf{BERT finetuned} & \textbf{98.0} & \textbf{87.5} & \textbf{92.4} & \textbf{93.6} \\
& GPT-3 & 39.6 \err{1.6}  & \textbf{82.7 \err{3.3}} & \textbf{90.5 \err{3.9}} & 45.1 \err{4.1} \\
& GPT-J & 27.0 \err{1.9} & 58.9 \err{3.1} & 58.4 \err{2.8} & 19.3 \err{1.9} \\

\bottomrule
\end{tabular}   

%% file: tabs/expansion_examples.tex
\begin{tabular}{lr|ll}
\toprule
Dataset &   \makecell[c]{Human\\score} & \makecell[l]{Example\\ CART ngram} &                                                                                    Added ngrams \\
\midrule
   \multirow{2}{*}{SST2} & \multirow{2}{*}{4.6\err{0.1}} &     good & fine, highly, solid, worthy, pleasing, satisfactory, outstanding, honorable, unwavering, valuable, ... \\
    &&      best & most remarkable, outstanding, superb, flawless, splendid, superlative, exceptional, impeccable, ... \\
     \multirow{2}{*}{RT} & \multirow{2}{*}{4.4\err{0.1}} &     dull & dreary, uninteresting, lackluster, listless, lifeless, uninspired, wearisome, drab, joylessly, ... \\
        &&       bad & unpleasant, dire, despicable, terrible, heinous, disgusting, vile, putrid, atrocious, nasty, poor, ... \\
\multirow{2}{*}{Emotion} & \multirow{2}{*}{4.4\err{0.2}} & miserable & gloomy, disillusioned, pathetic, doomed, agonized, despairing, pointless, despondent, ... \\
        &&     sorry & embarrassed, sorrowful, remorseful, excuse, unsatisfied, guilt, regretful, forgive, apologies, ... \\
\multirow{2}{*}{FPB} & \multirow{2}{*}{4.2\err{0.2}} & increased &                                                                                widened, consolidated \\
        &&      fell &                                slipped, slumped, diminished, plunged, dropped, weakened, lost ground \\        
\bottomrule
\end{tabular}

%% file: _main.bbl

%% file: appendix_gam.tex


\section{\methods}
\label{sec:appendix}

\begin{table}[ht]
    \centering
    \footnotesize
    \caption{Table of pre-trained models with unique huggingface identifiers.
    All models are used through huggingface~\citep{wolf2019huggingface},
    and linear/tree baselines are fit with scikit-learn~\citep{pedregosa2011scikit} and imodels~\cite{singh2021imodels}.}
    \input{tabs/tab_model_checkpoints}
    \label{tab:pretrained_models}
\end{table}

\begin{figure}[ht]
    \centering
    \includegraphics[width=0.6\textwidth]{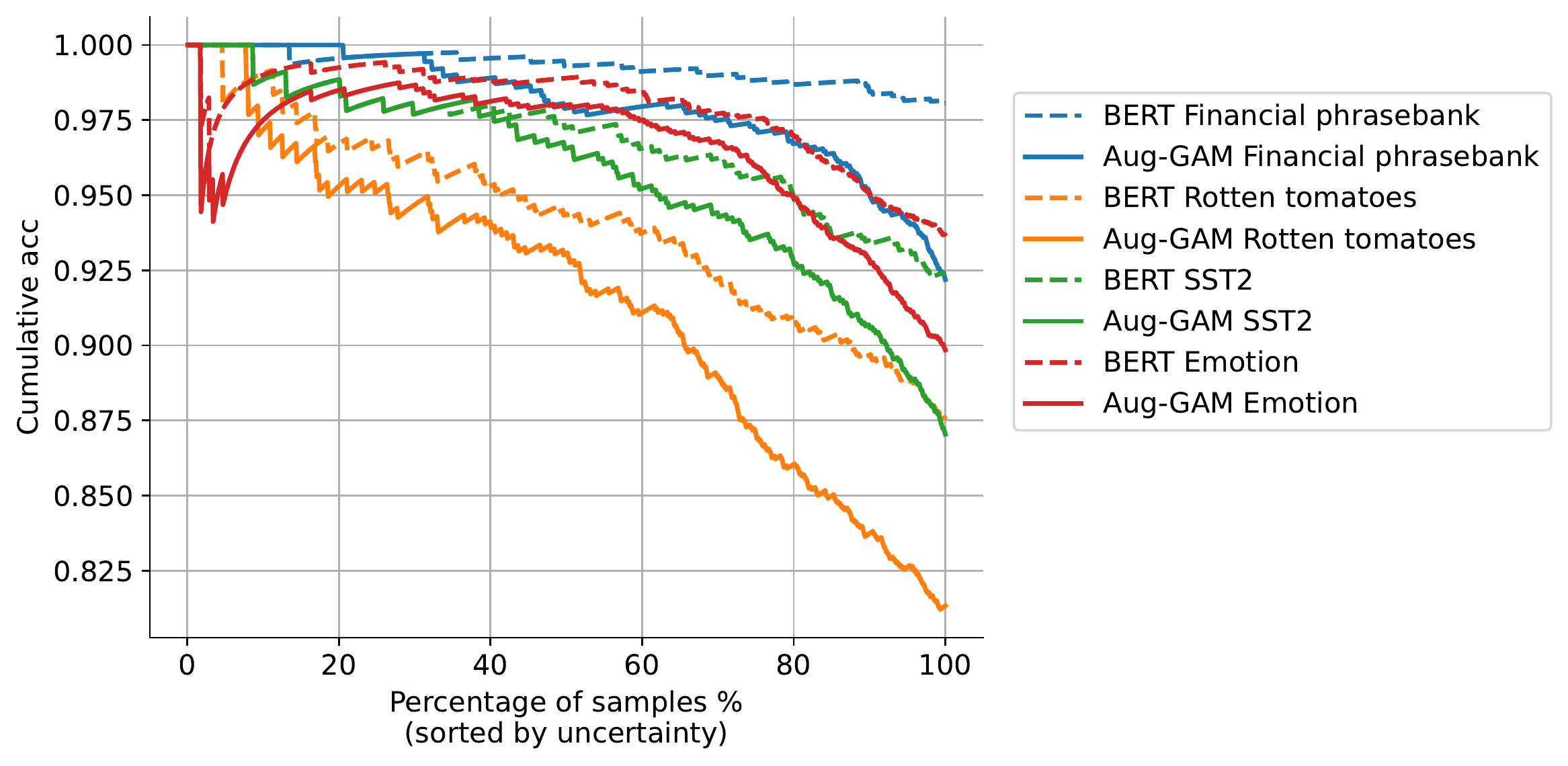}
    \caption{Model performance decreases with increasing model uncertainty. Cumulative validation accuracy decreases as more uncertain samples (based on the model's predicted probability) are added.}
    \label{fig:acc_calibration}
\end{figure}

\paragraph{Varying \methods settings} By default (\cref{tab:variation_results_full}), we use the final embedding layer of the model (and average it over the sequence length to get a fixed size vector), but \cref{tab:variation_results_full} also shows results using the \textit{pooler output} layer of the BERT model.
The choice of layer (i.e. final embedding layer versus pooler output) does not seem to make a large difference in the final performance results.
\cref{tab:variation_results_full} also shows one variation of the model (\textit{BERT finetuned (noun chunks)}) where rather than training on all ngrams, the model is fit to only noun-phrases extracted by spaCy's dependency parser~\citep{spacy2}.
This results in a performance drop across the datasets, suggesting that these noun-phrases alone are insufficient to perform the classification task.
We also run an experiment where we extract embeddings using Instructor (\cite{su2022one}, \texttt{hkunlp/instructor-xl}), which allows giving a contextual prompt for each dataset.

\paragraph{Evaluating zero-shot accuracy with language models}
To measure generalization ability, we evaluate explanations based on accuracy as a prompt for other models.
Accuracy is computed following \cite{brown2020language,raffel2020exploring}: using exact matching with beam search, a beam width of 4, and a length penalty of $\alpha = 0.6$. 
For sentiment evaluation, we use each prompt with the template \textit{Input: ``\$\{input\}''\{prompt\}}.\footnote{In initial experiments, we find that performance drops significantly when learning a prompt that comes \textit{before} the input.} We use \textit{positive} and \textit{negative} as positive and negative labels and require the LLM to rank the two options. Human-written prompts are adapted to this template from open-source prompts available through PromptSource \cite{bach2022promptsource}. 

\begin{table}[ht]
    \centering
    \small
    \footnotesize
    \caption{Generalization accuracy varies depending on the model used to extract embeddings.
    Finetuning the embedding model improves \methods performance, using a BERT model seems to outperform a DistilBERT model, and the layer used to extract embeddings does not have too large an effect. Top two methods are bolded in each column.\vspace{5pt}
    }
    \renewcommand{\arraystretch}{2}
    \input{tabs/variation_results_full}
    \label{tab:variation_results_full}
\end{table}

\subsection{Test-time tradeoffs between accuracy and interpretability/speed}
\label{subsec:efficiency_tradeoffs_gam}

\begin{figure}[ht]
    \centering
    \begin{overpic}[width=0.45\textwidth,tics=10]{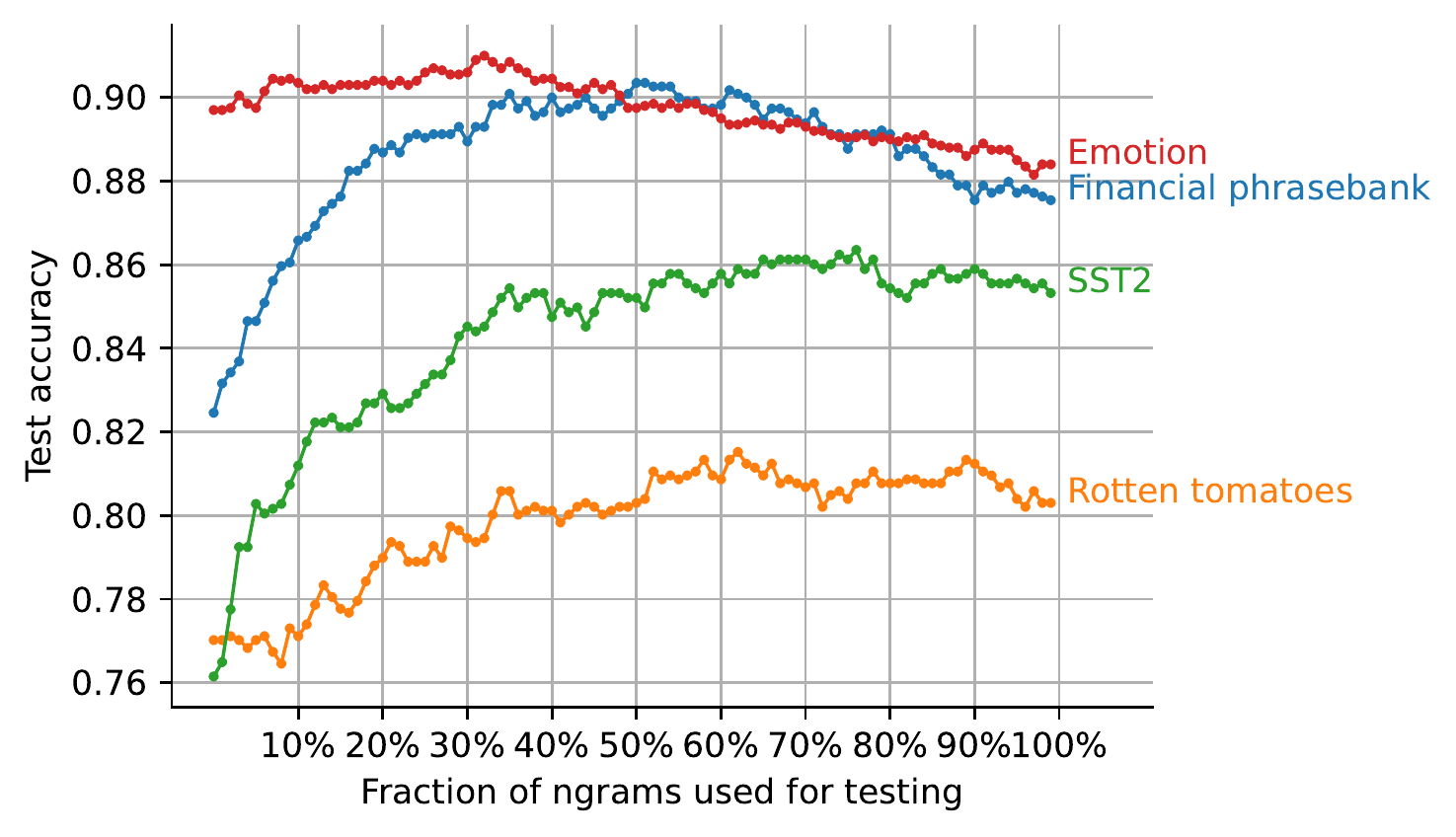}
        \put (1,55) {\large \textbf{A}}
    \end{overpic}
    \begin{overpic}[width=0.45\textwidth,tics=10]{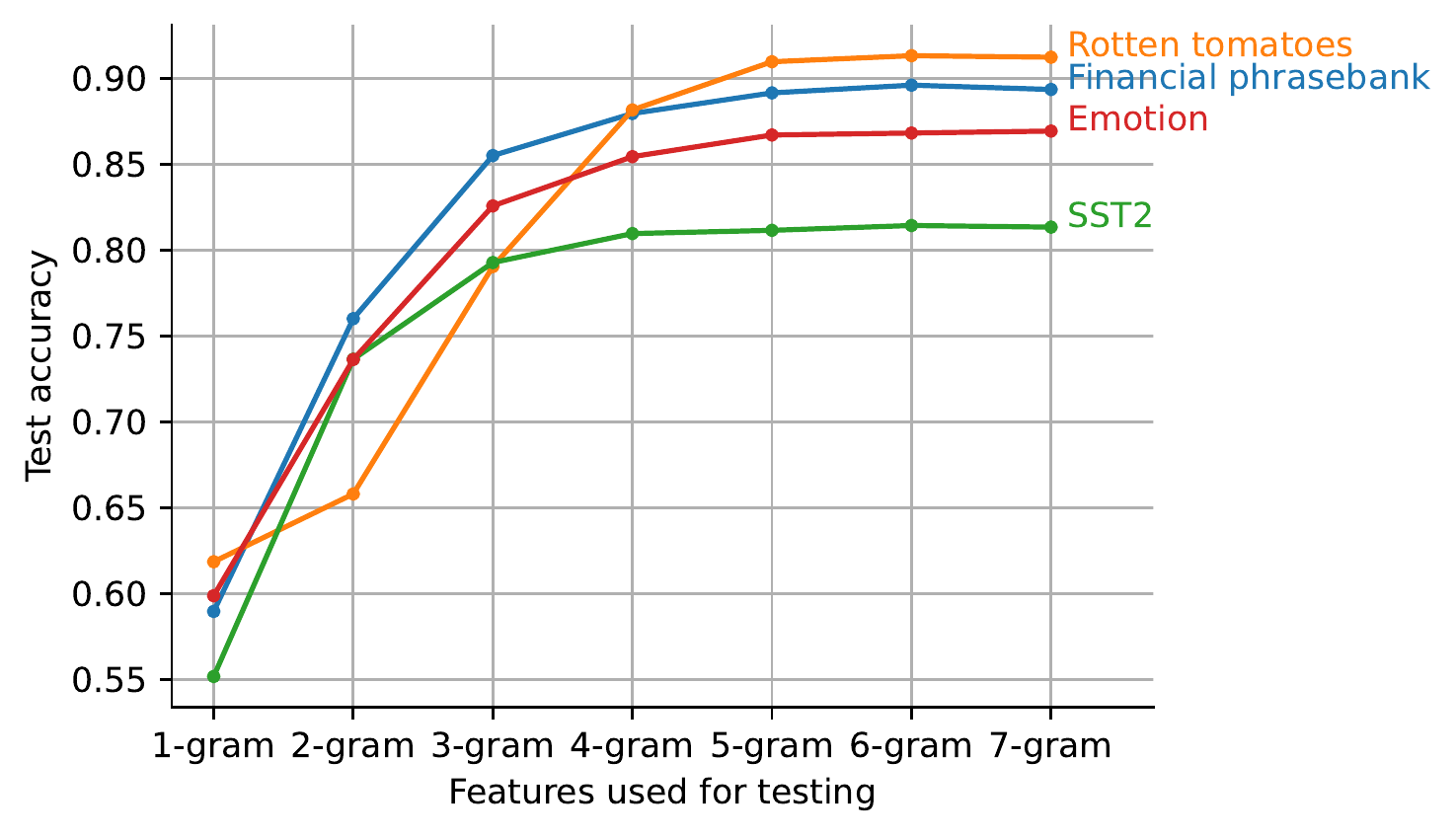}
        \put (1,55) {\large \textbf{B}}
    \end{overpic}
    \caption{\methods performance when varying the ngrams used for \textit{testing}.
    \textbf{(A)} Performance when removing the smallest coefficients from an \methods model.
    \textbf{(B)} Performance when varying the order of ngrams used for testing.
    }
    \label{fig:vary_ngrams_single}
\end{figure}

The ability to effectively generalize to unseen tokens in \cref{fig:ngram_contribution}C/D raises the question of whether one can vary the order of ngrams used \textit{at test-time}, to get a tradeoff between accuracy and interpretability (i.e. how many features are used to make a prediction).
Depending on the relative importance of accuracy and interpretability for a given problem, one may select to use a different number of features for testing.
\cref{fig:vary_ngrams_single} suggests that this is feasible.

\cref{fig:vary_ngrams_single}A shows the prediction performance when compressing the \methods model (fit using 4-grams and finetuned BERT) by setting the coefficients with the smallest magnitudee to zero.
Some models require only a few coefficients to perform well and some models (e.g. the \textit{Emotion} and \textit{Financial phrasebank} models) predict more accurately when using less than 50\% of the original coefficients.
\cref{fig:vary_ngrams_single}B it shows the accuracy of the same models in \cref{fig:vary_ngrams_single}A, as the order of ngrams used \textit{only for testing} is varied.
As the number of features used for testing increases, the performance tends to increase but interpretations become more difficult.

\begin{figure}[ht]
    \centering
    \includegraphics[width=0.35\textwidth]{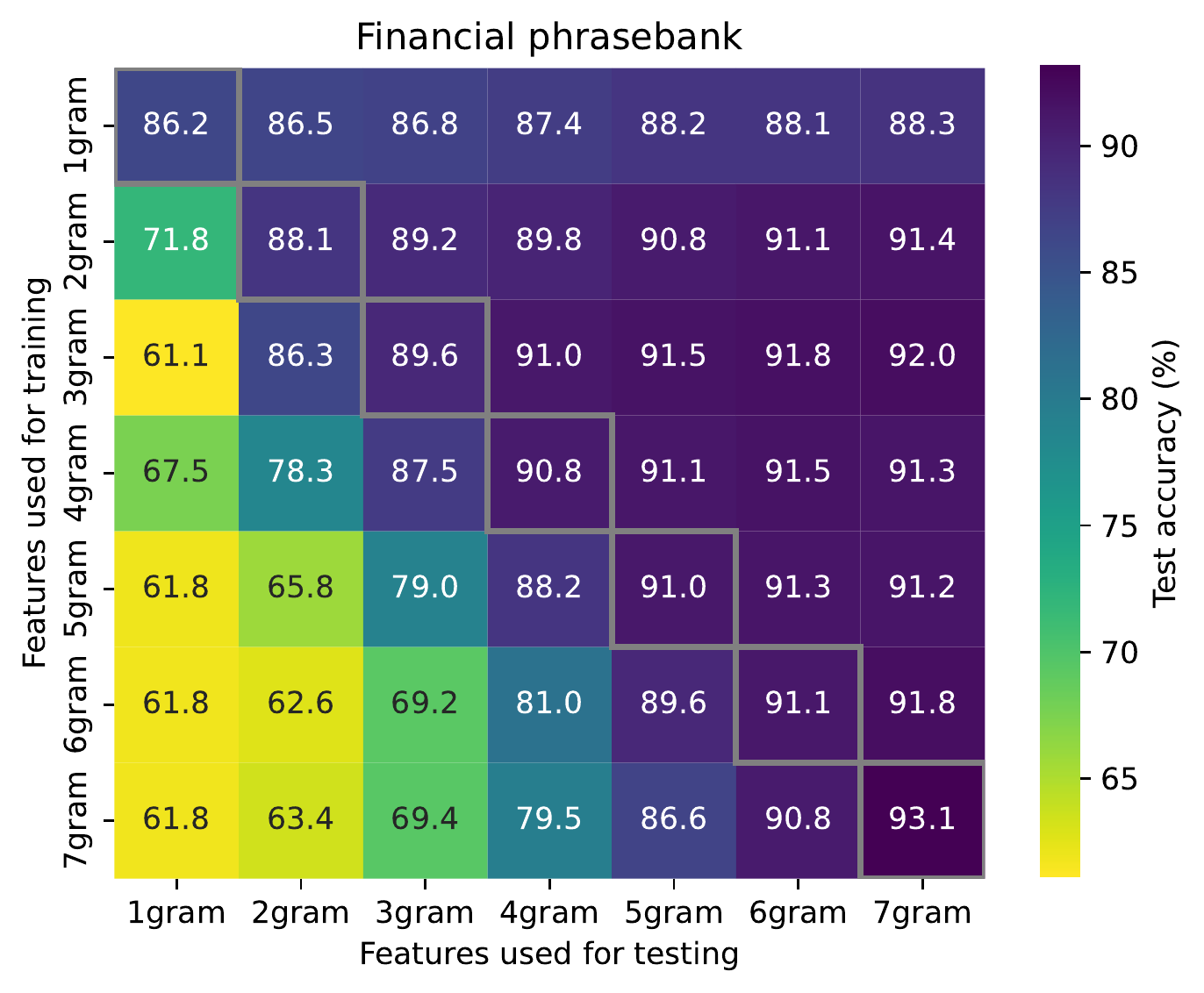}
    \includegraphics[width=0.35\textwidth]{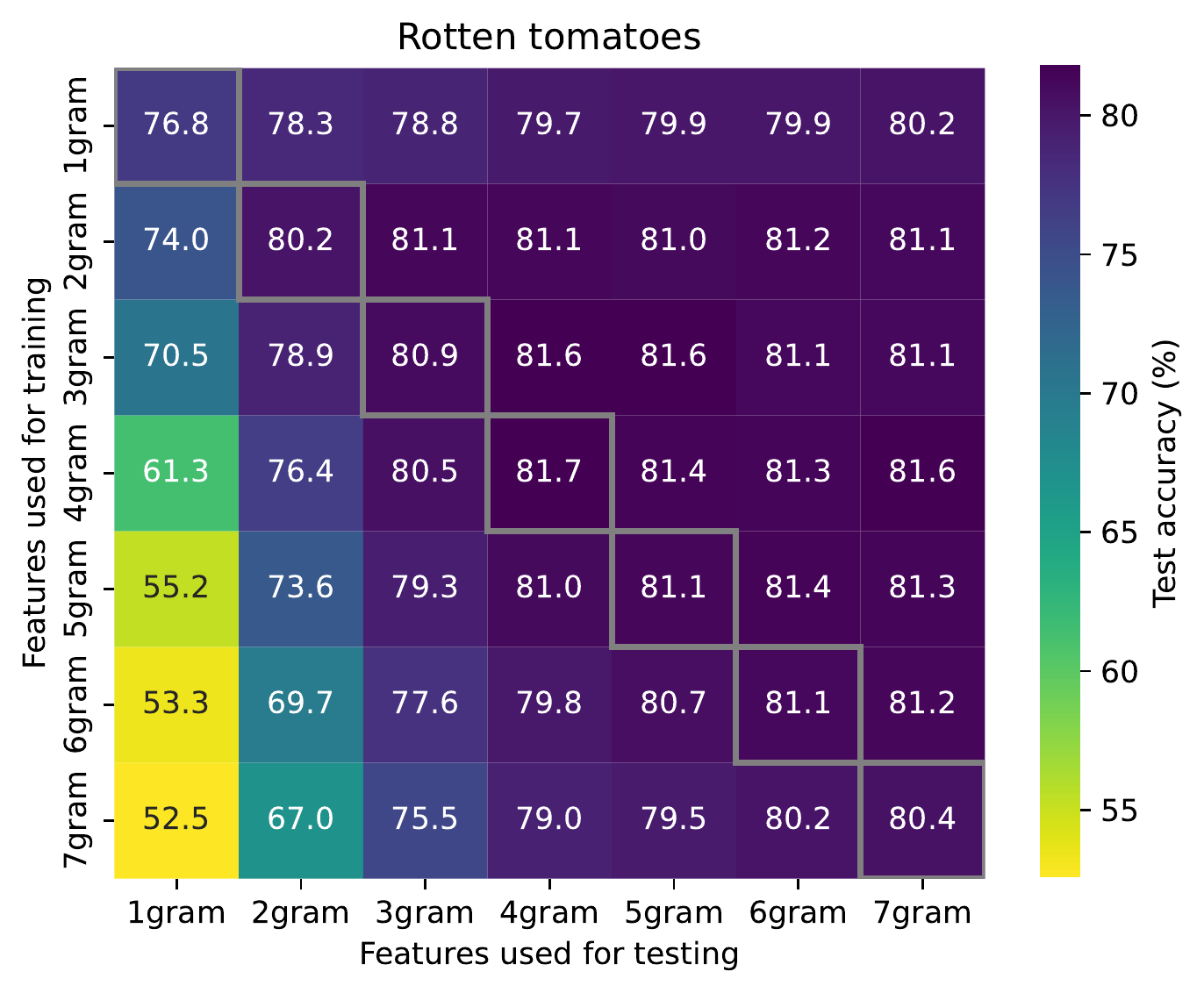}
    \includegraphics[width=0.35\textwidth]{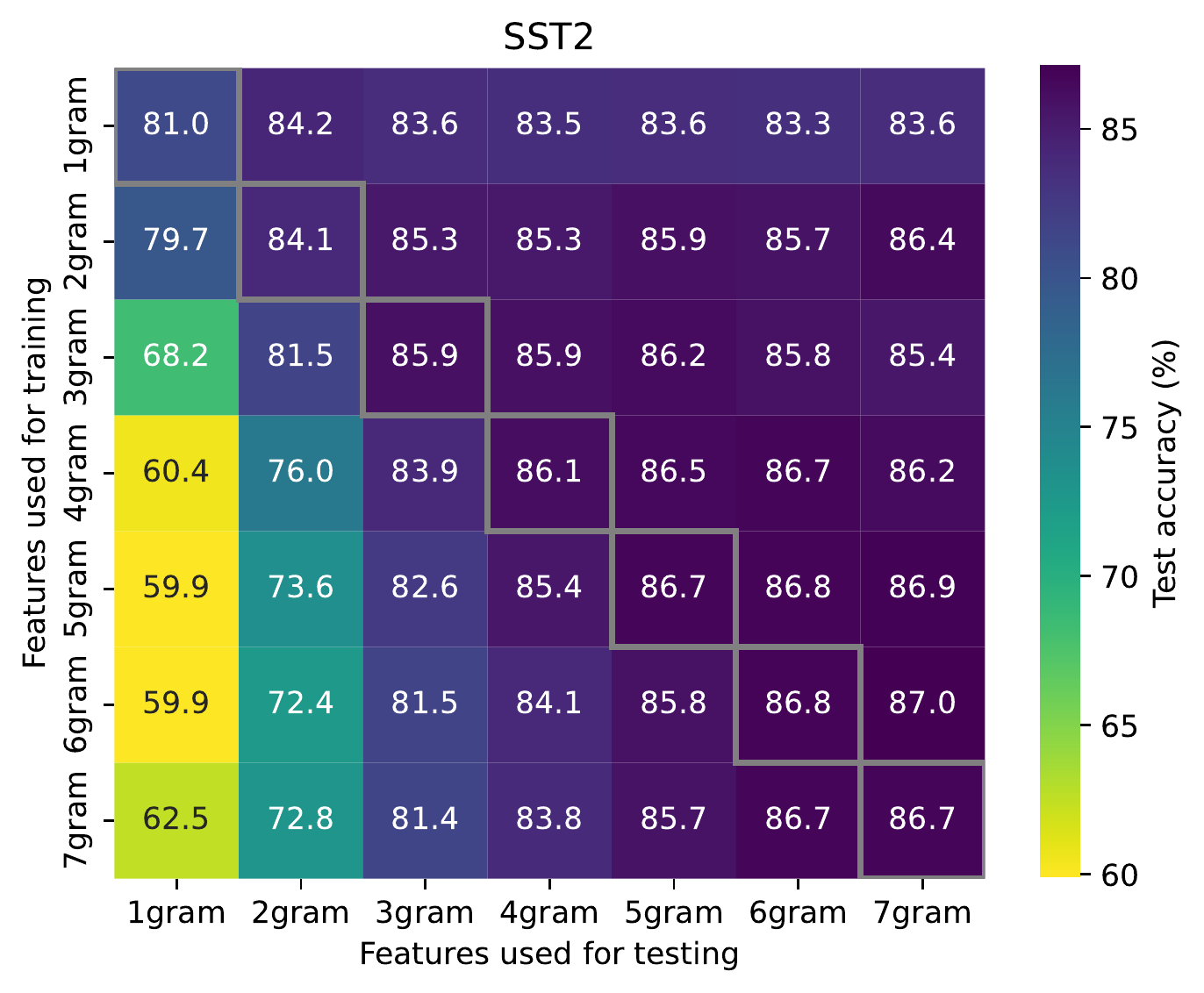}
    \includegraphics[width=0.35\textwidth]{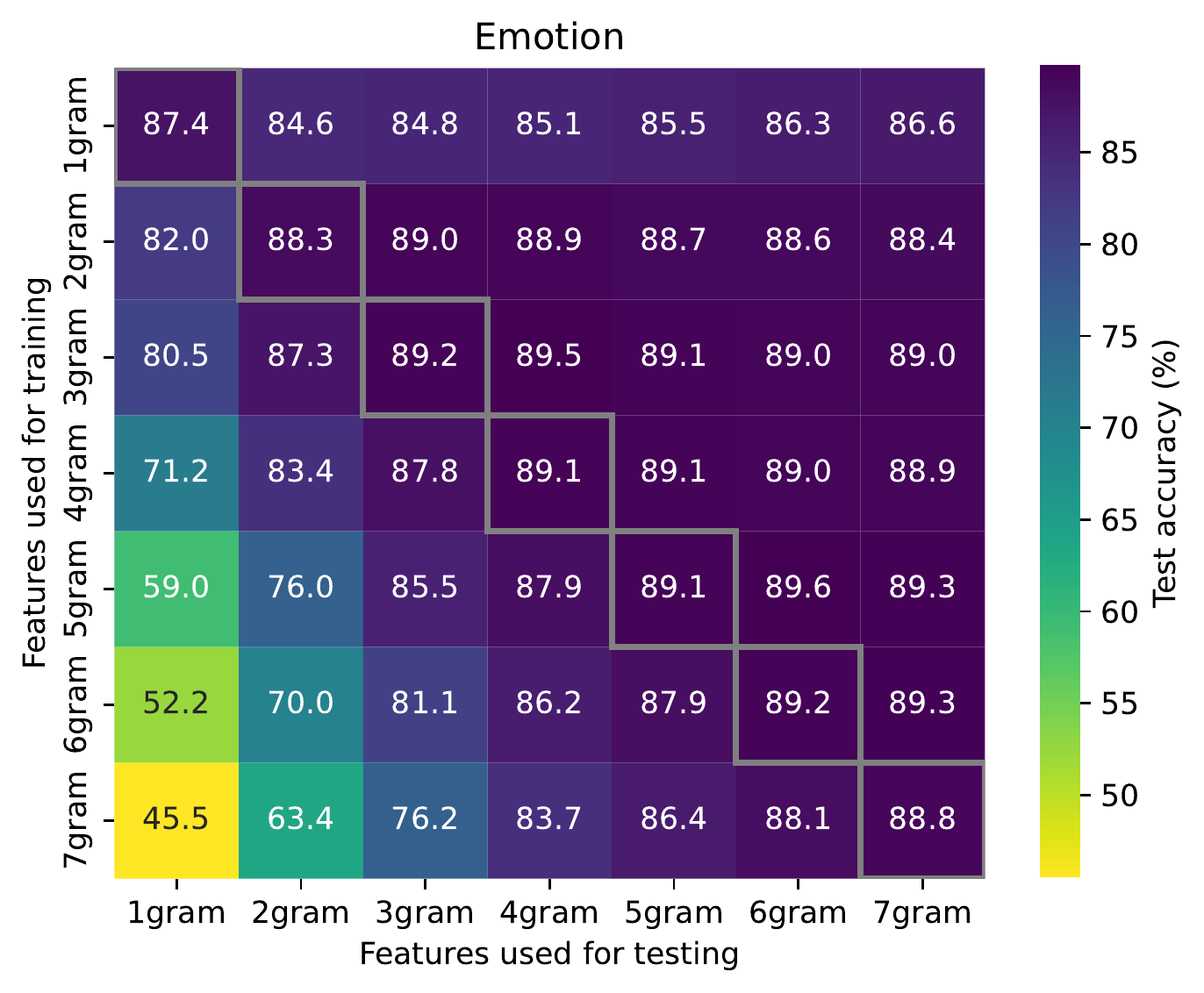}
    \caption{Varying the order of ngrams used for training and testing across each of the five datasets in \cref{tab:datasets_ovw}.
    Some models (i.e. rows) perform reasonably well as the order of ngrams used for testing is varied, potentially enabling a test-time tradeoff between accuracy and interpretability.
    Generally, using higher-order ngrams during testing improves performance and testing with less ngrams than used for training hurts performance considerably.
    }
    \label{fig:vary_ngrams_test}
\end{figure}

\cref{fig:vary_ngrams_test} characterizes the full tradeoff between the number of ngrams used for fitting versus testing for all datasets.
Generally, the best performance is achieved when the same number of ngrams is used for training and testing (the diagonal).
Performance tends to degrade significantly when fewer ngrams are used for testing than training (lower-left).

\subsection{Comparison with post-hoc feature importance}
\label{subsec:posthoc_comparison}

The coefficients learned by \methods often differ from importances assigned by post-hoc feature-importance methods.
\methods learns a single coefficient for each ngram across the dataset, allowing for auditing/editing the model with visualizations such as \cref{fig:ngram_contribution}.
In contrast, popular methods for post-hoc feature importance, such as LIME~\cite{ribeiro2016should} and SHAP~\cite{lundberg2016unexpected} yield importance scores that vary based on the context in each input.
This can be useful for debugging complex nonlinear models, but these scores (i) are approximations, (ii) must summarize nonlinear feature interactions, and (iii) vary across predictions, making transparent models preferable whenever possible.

\begin{figure}[th]
    \centering
    \includegraphics[width=0.5\columnwidth]{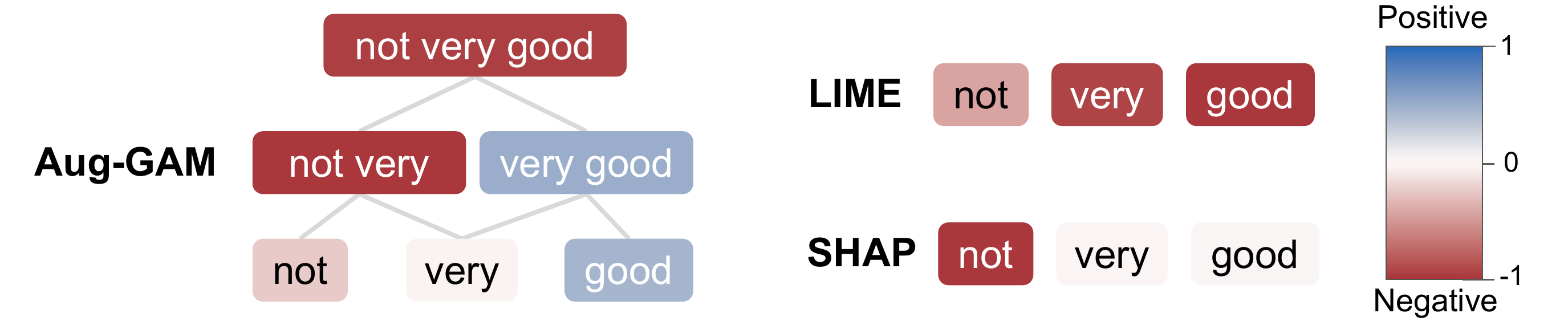}
    \caption{Comparing \methods ngram coefficients (left) to word-level feature importances from posthoc methods (right): LIME and SHAP.}
    \label{fig:posthoc_feat_importances}
\end{figure}

\begin{figure}[ht]
    \centering
    \includegraphics[width=0.7\textwidth]{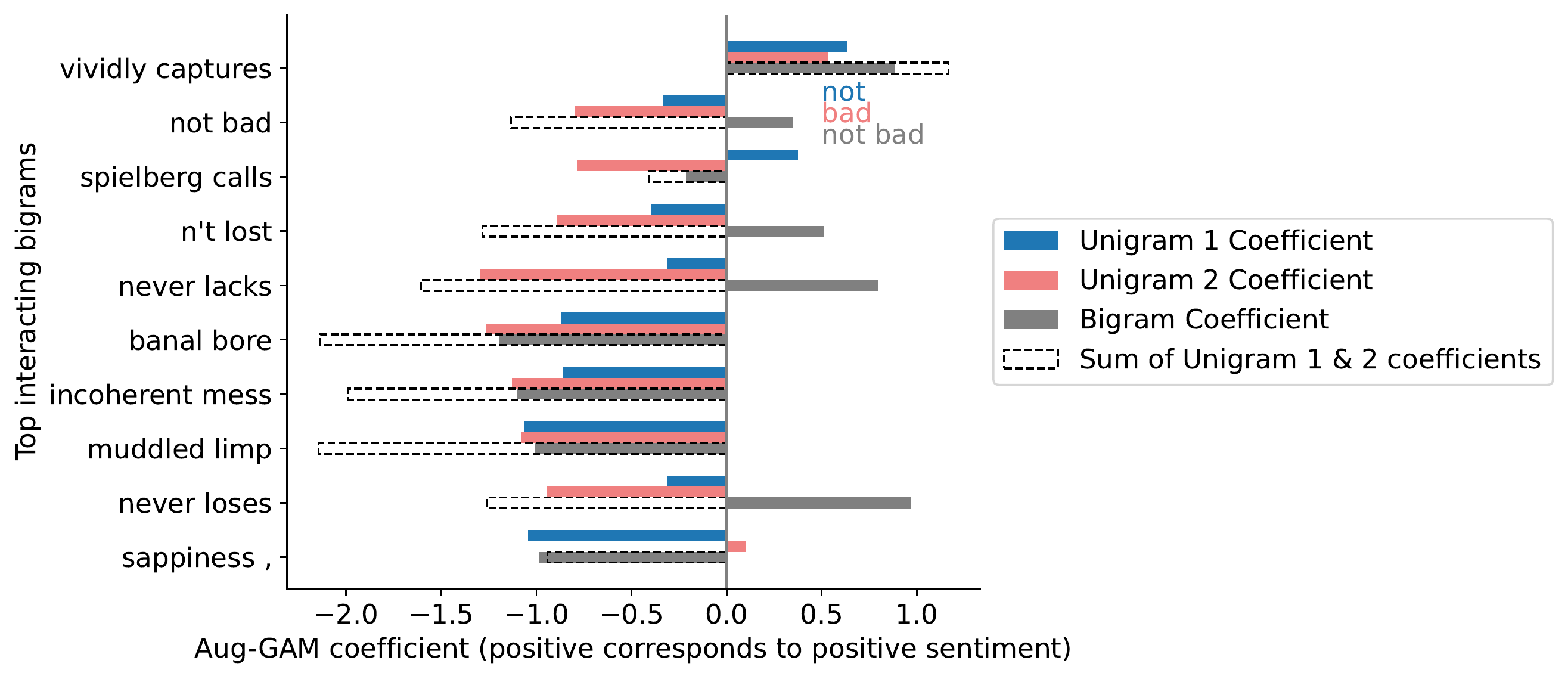}
    \caption{\methods accurately learns interactions rather than simply summing the contributions of individual unigrams.
    }
    \label{fig:bert_bow_comparison}
\end{figure}

\cref{fig:posthoc_feat_importances} shows an example of the \methods coefficients for the SST2 model from \cref{fig:ngram_contribution} for different ngrams  when making a prediction for the phrase \textit{not very good}.
While \methods yields scores for each subphrase that match human judgement (as seen in \cref{fig:ngram_contribution}B/D), posthoc feature importance methods summarize the interactions between different ngrams into individual words, potentially making interpretation difficult.
Scores are rescaled to be between -1 and 1 to make them comparable.
See \methods scores for many top-interacting phrases in \cref{fig:bert_bow_comparison}.

\paragraph{Summing embeddings meaningfully captures interactions}
One potential concern with the \methods model is that it may fail to learn interactions since it simply sums the embeddings of individual ngrams, and the language model extractor may not sufficiently capture interactions in its embedding space.
To investigate this concern, we first identify bigrams that involve interaction by fitting a unigram bag-of-words model and a bigram bag-of-ngrams model to \textit{SST2}.
We then use these two models to select the 10 bigrams for which the bigram coefficient is farthest from the sum of the coefficients for each unigram.

\cref{fig:bert_bow_comparison} shows the resulting bigrams containing interactions.
For each bigram, it shows the \methods learned coefficient (i.e. the contribution to the prediction $w^T \phi(x_i)$) for the bigram (gray bar) along with each of its constituent unigrams (blue and orange bars).
It is clear that the bigram coefficient is not the simple naive sum of the unigram coefficients (dashed black bar), and the learned coefficients make intuitive sense, suggesting that this \methods model has successfully learned interactions.

%% file: tabs/tab_model_checkpoints.tex
\begin{tabular}{ll}
    \toprule
    & BERT \\
    \midrule
    Base (no finetuning) & \tin{bert-base-uncased}~\citep{devlin2018bert} \\
    Emotion & \tin{nateraw/bert-base-uncased-emotion}\\
    Financial phrasebank & \tin{ahmedrachid/FinancialBERT-Sentiment-Analysis}~\citep{finbert} \\
     Rotten tomatoes & \tin{textattack/bert-base-uncased-rotten\_tomatoes}~\citep{morris2020textattack}\\
    SST2 & \tin{textattack/bert-base-uncased-SST-2}~\citep{morris2020textattack}\\
    \toprule
     & DistilBERT \\
    \midrule
    Base (no finetuning) & \tin{distilbert-base-uncased}~\citep{sanh2019distilbert}\\
    Emotion & \tin{aatmasidha/distilbert-base-uncased-finetuned-emotion}\\
    Financial phrasebank & \tin{yseop/distilbert-base-financial-relation-extraction}~\citep{akl2021yseop}\\
     Rotten tomatoes & \tin{textattack/distilbert-base-uncased-rotten-tomatoes}~\citep{morris2020textattack}\\
    SST2 & \tin{distilbert-base-uncased-finetuned-sst-2-english}\\
    \toprule
    & RoBERTa~\citep{liu2019roberta} \\
    \midrule
    Emotion & \tin{bhadresh-savani/roberta-base-emotion}\\
    Financial phrasebank & \tin{abhilash1910/financial\_roberta}\\
     Rotten tomatoes & \tin{textattack/roberta-base-rotten-tomatoes}~\citep{morris2020textattack}\\
    SST2 & \tin{textattack/roberta-base-SST-2}~\citep{morris2020textattack} \\
     \bottomrule
\end{tabular}

%% file: tabs/variation_results_full.tex
\begin{tabular}{lrrrr}
\toprule
{} & Financial phrasebank &      Rotten tomatoes &                 SST2  &              Emotion \\ 
\midrule
\makecell[l]{BERT finetuned}                  &  \textbf{92.8} \err{0.37} &  \textbf{81.6} \err{0.05} &   86.9 \err{0.1} &  \textbf{89.5} \err{0.03} \\ 
\makecell[l]{BERT finetuned\\(pooler output)}&  \textbf{93.5} \err{0.05} &  81.3 \err{0.13} &  \textbf{87.8} \err{0.21}  &  \textbf{89.8} \err{0.07} \\ 
\makecell[l]{BERT finetuned\\(noun chunks)}   &  87.9 \err{0.08} &  79.7 \err{0.45} &  84.1 \err{0.14} &   87.1 \err{0.2} \\ 
BERT                                       &  84.1 \err{0.08} &  78.1 \err{0.16} &  82.8 \err{0.27} &  67.1 \err{0.06}\\ 
\makecell[l]{BERT\\(pooler output)}           &  82.7 \err{0.28} &  78.5 \err{0.03} &  80.7 \err{0.11} &  58.0 \err{0.29} \\ 
DistilBERT finetuned                      &  85.8 \err{0.34} &  78.5 \err{0.34} &  81.7 \err{0.07}  &  68.8 \err{0.11} \\ 
DistilBERT                                 &  81.7 \err{0.34} &  79.8 \err{0.08} &   86.8 \err{0.1}  &  87.5 \err{0.11} \\ 
RoBERTa finetuned                                 & 77.8 \err{0.31}	 &  \textbf{83.6 }\err{0.03}	 &   \textbf{89.1} \err{0.24}	  &  88.5 \err{0.19}		 \\ 
Instructor prompted & 76.1 & 80.0 & 84.7 & 71.0 \\
\bottomrule
\end{tabular}

%% file: appendix_tree.tex
\FloatBarrier
\section{\methodts}
\label{sec:app_tree}

\renewcommand{\thealgorithm}{B\arabic{section}} 

\begin{algorithm}[t]
  \caption{\methodts algorithm for fitting a single split.}
  \label{alg:method}
  \scriptsize
\begin{algorithmic}[1]
  \State {\bfseries Split-\methodt}(\var X, \var y, \var{LLM}):  \\
  \vspace{5pt}
  \CommentMain{Add original CART keyphrase}
  \State \var{keyphrase} = \text{split\_CART}(\var X, \var y)  
  \State \var{keyphrases\_expanded} = LLM(\text{``Generate similar keyphrases to ''} + \var{keyphrase})
  \State \var{keyphrases\_running} = [\var{keyphrase}]
  \State{\var{impurity\_decrease\_best}} = \text{calc\_impurity\_decrease}(\var X, \var y, \var{keyphrases\_running})\\
  \vspace{5pt}
  \CommentMain{Try adding new keyphrases}
  \For{}\hspace{-3pt}\var{k} in \var{keyphrases\_expanded}:
    \State{\var{keyphrases\_running}.push(\var{k})}
    \State{\var{impurity\_decrease\_new}} = \text{calc\_impurity\_decrease}(\var X, \var y, \var{keyphrases\_running})
    \If{\var{impurity\_decrease\_new} $<$ \var{impurity\_decrease\_best}} 
        \State{\var{keyphrases\_running}.pop()}
    \EndIf
  \EndFor\\
  \vspace{1pt}
    \Return \var{keyphrases\_running}
\end{algorithmic}
\end{algorithm}

\begin{table}[t]
    \centering
    \scriptsize
    \caption{Metadata on keyphrase expansions.
    Results are averaged over keyphrases found in the 4 text-classification datasets in \cref{tab:datasets_ovw} when fitting a 40-tree bagging ensemble.
    The LLM is queried for 100 expansion candidates, but due to imperfect LLM generations, only 91.6 candidates are generated on average.
    After deduplication (converting to lowercase, removing whitespaces, etc.), only 83.3 candidates remain.
    Screening removes almost all candidates, leaving only 0.8 candidates on average.
    }
    \input{tabs/expansions_metadata}
    \label{tab:expansion_metadata}
\end{table}


\subsection{\methodts variations}
\label{subsec:variations_tree}
\cref{tab:variations_tree} explores different variations of \methodt.
The top row shows learning a single tree with \methodts using its default parameters, achieving the best performance across the datasets.
\cref{tab:variations_tree} shows results for different algorithmic choices, such as replacing the generic prompt with a dataset-specific one (\textit{\methodts (Contextual prompt)}), and searching for new keyphrases using 5 CART features instead of one (\textit{\methodts (5 CART features)}).
We also consider preprocessing the data differently, using \textit{Stemming} (with the Porter Stemmer) or using \textit{Trigrams}, rather than bigrams.

One major variation we study is using LLM embeddings to find keyphrases, rather than querying via a prompt (\textit{\methodts (Embeddings}).
Specifically, we consider expanding keywords by finding the keyphrases that are closest in embedding space (measured by euclidean distance) to the original keyphrase.
This option may be desirable computationally, as it may require a smaller LLM to compute effective embeddings (e.g. BERT~\cite{devlin2018bert}) compared to a larger LLM required to directly generate relevant keyphrases (e.g. GPT3~\cite{brown2020language}).
However, finding closest embeddings requires making more calls to the LLM, as embeddings must be calculated and compared across all ngrams in $X_{\text{text}}$.

\begin{table}[t]
    \centering
    \scriptsize
    \caption{Performance (ROC AUC) for variations of \methodt. Values are averaged over 3 random dataset splits; error bars are standard error of the mean (many are within the points).}
    \input{tabs/variations_tree.tex}
    \label{tab:variations_tree}
\end{table}

\begin{table}[t]
    \centering
    \scriptsize
    \caption{Performance (Accuracy) for \methodt and \methodts Ensemble. Values are averaged over 3 random dataset splits; error bars are standard error of the mean (many are within the points).
    *\textit{Emotion} and \textit{Financial phrasebank} results are not directly comparable to \cref{tab:best_results}, as they have been  modified for binary classification.
    }
    \input{tabs/variations_tree_acc.tex}
    \label{tab:variations_tree_acc}
\end{table}

%% file: tabs/expansions_metadata.tex
\begin{tabular}{rrr}
\toprule
\makecell{\# Expansion candidates\\(Before deduplication)} & \# Expansion candidates & \makecell{\# Expansions\\(After screening)}\\
\midrule
                                     91.6\err{0.7} &          83.3\err{0.8} &                               0.8\err{0.1}\\
\bottomrule
\end{tabular}

%% file: tabs/variations_tree.tex
\begin{tabular}{lllll}
\toprule
 & Emotion & Financial phrasebank & Rotten tomatoes & SST2 \\
\midrule
\textbf{Aug-Tree} & \textbf{0.680 \err{0.029}} & \textbf{0.825 \err{0.006}} & \textbf{0.622 \err{0.007}} & \textbf{0.673 \err{0.008}} \\
Aug-Tree (Embeddings) & 0.599 \err{0.008} & 0.776 \err{0.018} & 0.600 \err{0.011} & 0.663 \err{0.002} \\
Aug-Tree (Contextual prompt) & 0.667 \err{0.011} & 0.820 \err{0.004} & 0.627 \err{0.008} & 0.669 \err{0.005} \\
Aug-Tree (5 CART features) & 0.711 \err{0.039} & 0.730 \err{0.026} & 0.608 \err{0.009} & 0.674 \err{0.003} \\
Aug-Tree (Stemming) & 0.640 \err{0.019} & 0.520 \err{0.016} & 0.625 \err{0.004} & 0.679 \err{0.005} \\
Aug-Tree (Trigrams) & 0.676 \err{0.030} & 0.826 \err{0.006} & 0.619 \err{0.010} & 0.669 \err{0.006} \\
\midrule
CART & 0.574 \err{0.002} & 0.775 \err{0.005} & 0.599 \err{0.005} & 0.636 \err{0.002} \\
ID3 & 0.573 \err{0.004} & 0.795 \err{0.010} & 0.589 \err{0.002} & 0.638 \err{0.009} \\
\bottomrule
\end{tabular}

%% file: tabs/variations_tree_acc.tex
\begin{tabular}{lllll}
\toprule
 & Emotion* & Financial phrasebank* & Rotten tomatoes & SST2 \\
\midrule
\methodts & 0.637 \err{0.045}&	0.818 \err{0.014}&	0.613 \err{0.009}	&0.571 \err{0.018}\\
\methodts Ensemble & 0.800 \err{0.008} &	0.848 \err{0.006}&	0.619 \err{0.004}	&0.614 \err{0.016}\\
\bottomrule
\end{tabular}

%% file: appendix_fmri.tex
\FloatBarrier
\section{fMRI experiment details}
\label{sec:fmri_supp}

This section gives more details on the fMRI experiment analyzed in \cref{sec:fMRI}; for more scientific details see the original study~\citep{lebel2022natural}.
\cref{sec:fMRI} analyzes data from one human subject (\texttt{UTS03}) in the original study,
as the subject listened to approximately hours of narrative speech from the Moth Radio Hour, which consists of short autobiographical stories.
The subject underwent fMRI scanning as they listened, yielding an fMRI volume brain scan consisting of 95,556 voxels roughly every two seconds.

The individual voxel models described in \cref{sec:fMRI} are each fit to 9,461 training points, each corresponding to a different time point (after accounting for various preprocessing steps, such as trimming the beginning and end of the sequence).
They are evaluated on 291 volumes which come from a narrative story that was not seen during training.

\begin{figure}[ht]
    \centering
    \includegraphics[width=0.8\textwidth]{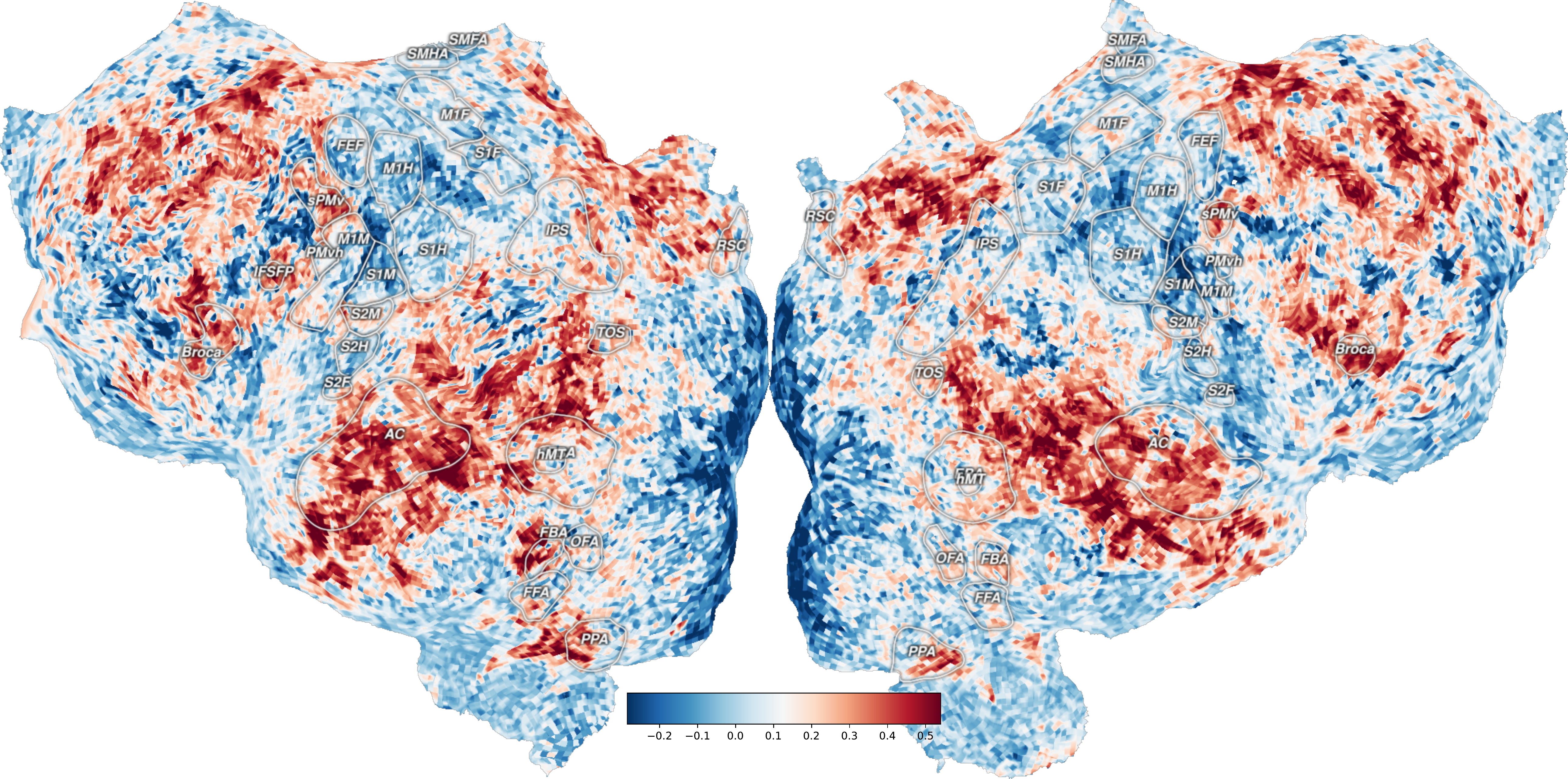}
    \caption{Generalization performance for individual-voxel models, measured by correlation between the predicted response and the measured response on the held-out test set. Some regions are very poorly predicted (blue), but many voxels can be predicted quite well (red).}
    \label{fig:flatmap_corr}
\end{figure}

\cref{fig:flatmap_corr} shows the generalization performance of the model for each voxel, measured by the correlation between the predicted response and the measured response.
\cref{fig:flatmap_diff} shows the performance difference between the \methods model and the BERT baseline.

\begin{table}[ht]
    \scriptsize
    \caption{fMRI prediction performance using different methods. Eng1000 is a linear word-embedding baseline similar to word2vec which has been used in the neuroscience literature~\cite{huth2016natural}.}
    \centering
    \input{tabs/fmri_supp.tex}
    \label{tab:fmri_supp_results}
\end{table}

%% file: tabs/fmri_supp.tex
\begin{tabular}{lccccc}
\toprule
            Model &  Order of ngram & $\rho$ &  \makecell{$\rho$\\(top 1\%)} &  \makecell{$\rho$\\(top 5\%)} \\
\midrule
          Eng1000 & 1 &   0.041 &                         0.529 &                         0.439 \\
            GloVe & 1&   0.044 &                         0.521 &                         0.426 \\
           BERT & 5 &   0.022 &                         0.386 &                         0.302 \\
          BERT & 10 &   0.035 &                         0.457 &                         0.365 \\
          BERT & 20 &   0.053 &                         0.524 &                         0.429 \\
 Aug-GAM (BERT) & 5 &   0.061 &                         0.583 &                         0.489 \\
\textbf{Aug-GAM (BERT)} & \textbf{10} &   \textbf{0.062}      &                         \textbf{0.586} &                         \textbf{0.489} \\
Aug-GAM (BERT) & 20 &   0.056 &                         0.566 &                         0.470 \\
\bottomrule
\end{tabular}